\documentclass[journal]{IEEEtran}
\usepackage[T1]{fontenc}
\usepackage[utf8]{inputenc}
\usepackage{pmboxdraw}
\usepackage{geometry}
\usepackage{ifpdf}
\usepackage{pmboxdraw}
\usepackage{cite}
\usepackage{booktabs, makecell}
\usepackage{multirow}
\usepackage{tabularx}
\AtBeginEnvironment{tabular}{\footnotesize}
\usepackage{graphicx}
\usepackage{textcomp}
\usepackage{color}
\definecolor{black25}{gray}{0.75}
\definecolor{gray8}{gray}{0.92}
\makeatletter
\long\def\rev#1{\begingroup\color{black}#1\endgroup}
\long\def\revision#1{\begingroup\color{black}#1\endgroup}

\makeatother
\usepackage[normalem]{ulem} 

\usepackage{amsmath,bm}
\usepackage{amsfonts,amssymb}

\usepackage[linesnumbered,ruled,vlined]{algorithm2e}

\usepackage{array}

\usepackage[font=footnotesize]{subcaption}

\usepackage{url}
\usepackage{hyperref}
\usepackage[nameinlink, capitalise]{cleveref}

\begin{document}

%
\title{A Self-Evolving Defect Detection Framework for Industrial Photovoltaic Systems}

%
%
%

\author{Haoyu He, 
        Yu Duan,
        Wenzhen Liu, 
        Hanyuan Hang,
        Boyu Qin,
        Qiantu Tuo,  
        Xiaoke Yang,  
        Rui Li
\thanks{This work was supported by EcoFlow R\&D Projects.  (\emph{Corresponding author}: Rui Li).}
\thanks{H. He, Y. Duan, W. Liu, H. Hang, Q. Tuo, X. Yang, and R. Li are with EcoFlow Inc, Shenzhen, China (eeairicky@gmail.com, {ricky.li}@ecoflow.com).}
\thanks{B. Qin is with School of Electrical Engineering, Xi’an Jiaotong University,
Xi’an, China (qinboyu@xjtu.edu.cn)}}

\maketitle

\begin{abstract}
Reliable photovoltaic (PV) power generation requires timely detection of module defects that may reduce energy yield, accelerate degradation, and increase lifecycle operation and maintenance costs during field operation.
Electroluminescence imaging has therefore been widely adopted for PV module inspection. 
However, automated defect detection in real operational environments remains challenging due to heterogeneous module geometries, low-resolution imaging conditions, subtle defect morphology, long-tailed defect distributions, and persistent data and label-space shifts introduced by evolving inspection and labeling processes.
These factors significantly limit the robustness and long-term maintainability of conventional deep-learning inspection pipelines.
To address these challenges, this paper proposes \textit{SEPDD}, a \textit{Self-Evolving Photovoltaic Defect Detection} framework designed for evolving industrial PV inspection scenarios.
SEPDD integrates automated pipeline optimization with a trigger-driven evolution mechanism that re-optimizes the entire training/evaluation pipeline when deployment indicators (retraining failure, label evolution, environmental change, periodic maintenance) fire, enabling deployment-time adaptation to distribution shifts and newly emerging defect patterns.
Experiments conducted on both a public PV defect benchmark and a private industrial dataset demonstrate the effectiveness of the proposed framework. 
Both datasets exhibit severe class imbalance and significant domain shift.
SEPDD \revision{attains competitive mAP50 of 91.5\% on the public dataset and 49.5\% on the private dataset in our evaluation, where it} surpasses the autonomous baseline by 14.9\% and human experts by 4.8\% on the public dataset, and by 4.9\% and 2.5\%, respectively, on the private dataset\revision{; these comparisons are limited to the two evaluated datasets and the studied baselines}.
By improving the robustness and adaptability of automated PV inspection, the proposed framework supports more reliable maintenance of distributed PV systems.
\end{abstract}



\section{Introduction}
\label{sec:intro}

\IEEEPARstart{P}{hotovoltaic} (PV) generation has become one of the fastest-growing renewable energy sources and plays an increasingly important role in modern power systems. Ensuring the long-term reliability and sustained energy yield of PV modules is essential for stable power generation and effective lifecycle operation and maintenance (O\&M) management. In practical deployments, defects and degradation mechanisms---microcracks, hot spots, gridline interruptions, black cores, potential-induced degradation---may emerge during long-term field operation and, if not detected in time, significantly reduce power output, accelerate degradation, and increase O\&M costs \cite{ozkalay2025three,tang2025understanding}. Efficient and reliable defect inspection is therefore a key component of PV system reliability management.

PV module inspection commonly relies on imaging-based diagnostics such as electroluminescence (EL), infrared, and visible-light imaging; among these, EL imaging is particularly effective at revealing internal electrical and structural defects within photovoltaic cells. Conventional EL inspection workflows often require manual interpretation, which is labor-intensive, subjective, and difficult to scale to large fleets. Deep-learning approaches for automated PV defect detection \cite{liu2023efficient,akram2025advancing} have therefore been explored, including Transformer-based architectures \cite{barnabe2023quantification}, multi-scale feature fusion \cite{chen2024polycrystalline}, and attention-enhanced object detection \cite{lang2024pv}.

Despite these advances, deploying learning-based inspection systems in real operational environments remains fundamentally challenging. First, PV inspection datasets are typically small and strongly long-tailed---a few dominant defect types account for most samples while many reliability-critical defects appear rarely \cite{liu2023efficient,akram2025advancing}---and standard supervised pipelines are easily dominated by head classes, yielding poor recall for rare but high-risk defects \cite{he2009learning,cui2019class,tan2020equalization,tang2025understanding}.
Second, PV defect patterns exhibit substantial morphological complexity and semantic ambiguity \cite{deitsch2019automatic,wang2022deep}: appearances vary across module designs, installation conditions, and imaging devices, so samples in the same category may show large intra-class variation while visually similar patterns may correspond to different defect types under low-resolution EL imaging. Defects on abnormal PV cells often resemble textured background in EL images, which makes them difficult to distinguish from normal patterns or impurities \cite{wang2022deep}.
Third, PV inspection environments are dynamic: variations in imaging devices, illumination, installation, and operational conditions introduce persistent distribution shifts that degrade models trained on static datasets \cite{delpero2023long,kumar2023assessment}. The defect taxonomy itself evolves as new defect patterns emerge, and conventional pipelines that assume a fixed label space require repeated manual annotation and full retraining when new categories appear, making them difficult to maintain.
These observations suggest that PV defect inspection should be treated as a long-term adaptive task rather than a one-time closed-set problem: effective systems must remain robust under small, imbalanced data, adapt to distribution shifts, and continuously incorporate emerging 
categories.

To address these challenges, this paper proposes \textit{SEPDD}, a self-evolving defect detection framework for industrial PV inspection. Unlike conventional pipelines that fix the training script and optimize only weights or hyperparameters, SEPDD treats the entire training/evaluation pipeline as the search target and re-optimizes it through trigger-driven evolution cycles, addressing long-tailed data, distribution shift, and dynamically expanding defect taxonomies at the pipeline level rather than the weight level.
Our main contributions are as follows:
\begin{itemize}
\item[\textit{(i)}]
We identify key challenges in automated PV defect inspection---limited training data, long-tailed defect distributions, evolving inspection workflows, and emerging defect categories---and propose \textit{SEPDD}, a deployment-oriented self-evolution framework that integrates indicator-triggered evolution cycles with automated pipeline optimization to address these challenges.
\item[\textit{(ii)}]
We provide a deployment-oriented self-evolution stack for PV EL inspection that combines indicator-triggered evolution cycles, a quality-gated per-node operator workflow producing evaluated training pipelines before graph expansion, and a top-$k$ search with a global journal \revision{whose summary channel propagates cross-branch experience}, extending autonomous ML agents such as AIDE with mechanisms aimed at stable industrial adaptation under long-tailed data, domain shift, and evolving labels.
\item[\textit{(iii)}]
Extensive experiments on both public and industrial EL datasets demonstrate that SEPDD \revision{substantially} improves rare-defect detection performance while maintaining real-time inference capability\revision{, within the scope of the studied datasets and baselines}.
\end{itemize}

The remainder is organized as follows.
\Cref{sec:problem} describes the industrial PV inspection setting and challenges; \cref{sec:methodology} presents SEPDD; \cref{sec:experiments} reports experimental results; \cref{sec:conclusion} concludes the paper.

\section{Industrial PV Defect Detection}
\label{sec:problem}

\begin{figure*}[!t]
\centering
\includegraphics[width=0.98\linewidth]{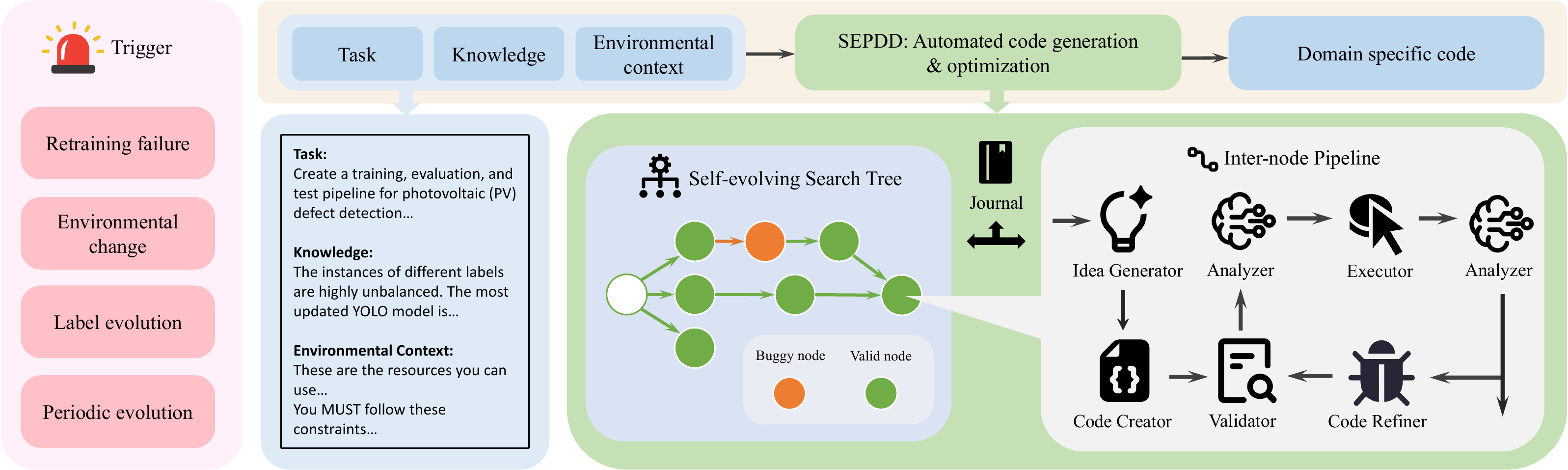}
\caption{\textbf{Overall architecture of the SEPDD framework.} Left: deployment triggers (retraining failure, label evolution, environmental change, periodic maintenance) start an evolution cycle. Center: the self-evolving search tree with \revision{a global journal records each pipeline exploration as a node and feeds summaries of the best-performing branches back into subsequent expansions.} Right: the per-node workflow \revision{(Algorithm~\ref{alg:node-workflow})} generates and refines a complete training pipeline before the node is admitted to inter-node expansion. Bottom: Task / Knowledge / Environmental Context supplies domain inputs to all operators.}
\label{fig:framework-overview}
\end{figure*}

\subsection{Problem Setting}

Reliable PV power generation critically depends on the
early identification of module defects before they propagate into
energy-reducing failure modes during field operation. Undetected
microcracks, metallization failures, dark regions, or localized
hotspots may expand under environmental stressors such as thermal
cycling, humidity–freeze, and mechanical loading, eventually leading
to irreversible degradation, energy-yield loss, and increased
O\&M costs over the module lifetime
\cite{aghaei2022review,kim2021review}. Accurate defect detection is
therefore essential for ensuring long-term PV system reliability and
stable energy yield.

\subsubsection{Industrial EL inspection setting}

EL imaging is widely used for PV module defect inspection because it reveals electrical inhomogeneities and structural defects within photovoltaic cells that are difficult to detect using visible-light or infrared imaging.
However, industrial EL inspection environments differ substantially from the controlled acquisition conditions of most public datasets \cite{aghaei2022review,deitsch2019automatic}: real-world EL images often exhibit heterogeneous module geometries, non-uniform spatial resolutions, acquisition noise, and evolving defect characteristics across imaging devices and production batches.
Industrial inspection datasets typically contain limited annotated samples and exhibit greater morphological diversity than public benchmarks such as \cite{deitsch2019automatic}, which significantly increases the difficulty of reliable automated defect detection.

\subsubsection{Data characteristics: limited annotated data and long-tailed distributions}

Industrial EL inspection datasets often contain limited annotated samples because EL imaging and expert annotation are expensive and time-consuming, providing insufficient statistical coverage of defect variations and increasing the risk of overfitting.
PV defect datasets also typically exhibit highly long-tailed distributions, where a few dominant defect categories account for most samples while reliability-critical defects appear only rarely \cite{zhang2021long}, and missing these rare defects may lead to significant reliability risks and long-term energy yield losses.

\subsection{Challenges}

Although deep-learning–based PV inspection methods have demonstrated
strong performance on curated academic datasets, deploying them in
real industrial environments introduces several fundamental challenges.

\subsubsection{Challenge 1: Complex defect morphology under low-resolution EL imaging}

PV defects originate from diverse mechanical, electrical, and
material degradation mechanisms \cite{aghaei2022review} and may exhibit widely varying shapes, scales, contrasts, and textures in EL imagery. Industrial inspection further increases this complexity due to heterogeneous panel geometries, lower imaging resolution, and acquisition noise, so samples in the same defect category may show large intra-class variation while visually similar patterns may correspond to different defect types.
Although modern object detection architectures and multi-scale feature representations have improved defect recognition performance \cite{deitsch2019automatic,wang2022deep}, these models are typically trained on relatively clean datasets and may not generalize well to industrial inspection scenarios.

\subsubsection{Challenge 2: Distribution shifts across operational environments}

Industrial PV inspection data are subject to persistent distribution shifts caused by variations in imaging devices, illumination conditions, module geometries, and production processes; the statistical characteristics of EL images may differ substantially from those of historical training data, and models trained on static datasets often experience significant performance degradation when deployed in new inspection environments.
Although domain adaptation techniques can partially mitigate such shifts by aligning feature distributions across domains \cite{wang2020deep}, maintaining stable detection performance under continuously changing conditions remains difficult.

\subsubsection{Challenge 3: Emerging defect categories and evolving label space}

Unlike academic benchmarks with fixed label taxonomies, the defect category space in industrial PV inspection is inherently dynamic.
As module technologies, materials, and manufacturing processes evolve, previously unseen defect types may gradually emerge; these new categories are typically rare and sparsely labeled in early stages but may pose significant risks to module reliability.
Conventional detection pipelines assume a fixed label space and require manual re-annotation and full model retraining when new defect categories appear. Continual learning (CL) techniques aim to address evolving tasks via incremental weight updates with anti-forgetting mechanisms \cite{parisi2019continual}, but in industrial PV inspection a new defect category typically requires re-optimizing the entire training recipe (sampling, augmentation, class weighting, backbone choice) rather than only fine-tuning weights, which CL methods do not address directly.

In summary, industrial PV defect inspection is characterized by small datasets, complex defect morphology, evolving data distributions, and emerging defect categories; these expose the limitations of static, closed-set detection pipelines and highlight the need for adaptive inspection systems.

\subsection{From Current Approaches to SEPDD}

AutoML and neural architecture search (NAS) reduce manual engineering by optimizing within a predefined space of architectures, operators, or hyperparameters over a fixed training script \cite{elsken2019neural}; autonomous ML engineering agents such as AIDE \cite{jiang2025aide} extend this paradigm by framing ML engineering as code optimization via LLM-driven tree search, demonstrating that strong pipelines can be discovered in code space rather than within a predefined space. Directly applying these methods to industrial PV EL inspection remains challenging due to severe class imbalance, limited training data, and evolving conditions; in practice, domain experts manually tune architectures, training strategies, and hyperparameters to obtain acceptable performance, but such tuning is effort-intensive and difficult to maintain as inspection conditions evolve.

We propose \textit{SEPDD}, a self-evolving framework that builds on the LLM-driven code-search paradigm and specializes it for long-term industrial PV EL inspection. \emph{vs.\ AutoML/NAS:} SEPDD searches over a program space---each candidate is a complete Python pipeline that may alter data handling, augmentation, sampling, optimization, model choice, and post-processing under task-specific constraints---so AutoML/NAS optimization within a predefined space is a constrained special case. \emph{vs.\ AIDE:} SEPDD differs in (i)~industrial objective and outputs (PV EL inspection under long-tailed defects, low-resolution morphology, evolving labels, and deployment constraints, exporting both pipeline code and the trained detector for redeployment when triggers fire); (ii)~a quality-gated per-node workflow so only executable, metric-bearing nodes become search parents---unlike AIDE, where a larger fraction of branches may not yield deployable code; and (iii)~top-$k$ inter-node search with a global journal \revision{whose summary channel propagates the experience of the best-performing branches across expansions}, replacing AIDE's simpler draft/debug/improve progression. \emph{vs.\ continual learning (CL):} SEPDD does not perform online incremental weight updates with explicit anti-forgetting mechanisms; ``self-evolving'' here refers to system-level evolution of the inspection pipeline, where deployment indicators trigger re-optimization of the full training/evaluation script and re-export of the detector. SEPDD is complementary to CL: any retention objective (e.g., replay, regularization) can be implemented inside an evolved pipeline. \Cref{fig:framework-overview} summarizes the architecture as a roadmap for \cref{sec:methodology}: triggers (left), self-evolving search with journal (center), and validated per-node pipeline generation (right).

\section{SEPDD: A Self-Evolving PV Defect Detection Framework}
\label{sec:methodology}

\subsection{Framework Overview}

As the architecture in \cref{fig:framework-overview},  deployment triggers
start an evolution cycle when retraining fails, labels change, the environment shifts, or periodic maintenance is due; the search tree and journal store successive pipeline explorations and their metrics; and the per-node workflow implements Algorithm~\ref{alg:node-workflow} to produce executable, evaluated code before inter-node expansion. Task / Knowledge / Environmental Context supplies domain inputs to all operators. SEPDD is an autonomous, trigger-driven framework that updates the detector when the pipeline can no longer maintain expected performance; for a given task, the solution is represented by source code that defines the ML training pipeline and produces the best-performing trained model.

\revision{\subsubsection{Quantitative trigger indicators}
\label{sec:trigger-indicators}
The framework tracks a set of indicators to determine when an evolution cycle is activated. A typical trigger is \emph{retraining failure}, where routine refreshment can no longer recover the desired performance; other triggers include \emph{label evolution} (new defect categories appear), \emph{operational/environmental changes} (changes in imaging conditions or deployment constraints), and \emph{periodic evolution} (scheduled to maintain model freshness).
In deployment, each of the four trigger families is monitored through quantitative indicators computed on the labeled QC stream. \emph{(i) Retraining failure:} a sliding-window mAP50 on the QC stream that persists below a floor of $\mu_{\mathrm{ref}} - \delta$ after a routine refresh (e.g., $\delta = 2$ percentage points over a 2-week window relative to the promoted reference $\mu_{\mathrm{ref}}$), or a recall drop concentrated on a subset of classes. \emph{(ii) Label evolution:} the share of QC images that fail a confidence filter and enter human review, together with the emergence rate of novel-class proposals per review batch; a sustained rise above baseline signals taxonomy drift. \emph{(iii) Environmental change:} statistical drift detectors (e.g., embedding-distance or KL divergence between feature histograms of recent and reference batches) exceeding a calibration threshold. \emph{(iv) Periodic evolution:} a calendar or device-mileage counter, which does not depend on data statistics. Indicator thresholds are calibrated on historical QC data so that the false-alarm rate of the statistical triggers stays low, and multiple statistical indicators must co-fire (or a single trigger persist across two evaluation windows) before an expensive cycle is started.
}

\begin{algorithm}[t]
\small
\caption{Workflow of one node (single expansion step).}
\label{alg:node-workflow}
\SetAlgoLined
\SetKwInOut{Input}{input}
\SetKwInOut{Output}{output}
\DontPrintSemicolon
\Input{parent node, SEPDD input, journal.}
\Output{new node}
\tcp{Code generation}
\textit{suggestions} $\gets$ \text{IdeaGenerator}(\textit{parent\_node})\;
\textit{code} $\gets \text{CodeCreator}(\textit{parent\_node, suggestions})$\;
\BlankLine
\tcp{Code refinement}
\Repeat{reach maximum debug depth}{
    \textit{syntax\_info}, \textit{exec\_output} $\gets \text{Validator}(\textit{code})$\;
    \textit{buggy}, \textit{analysis} $\gets$ \text{Analyzer}(\textit{code, syntax\_info, exec\_output})\;
    \If{not \textit{buggy}}{
        \textit{exec\_output} $\gets \text{Executor}(\textit{code})$\;
        \textit{buggy}, \textit{analysis} $\gets$ \text{Analyzer}(\textit{code, syntax\_info, exec\_output})\;
        \If{not \textit{buggy}}{
            break\;
        }
    }
    \textit{code} $\gets$ \text{CodeRefiner}(\textit{code, exec\_output, analysis})\;
}
\end{algorithm}

\revision{ \subsubsection{Source of domain knowledge}
\label{sec:domain-knowledge}
A key characteristic of SEPDD is that it is knowledge-based. Three sources feed the LLM-driven operators 
: \textbf{(A) human experts} contribute defect definitions and inspection priors (defect taxonomy, ``\texttt{star\_crack} is rare,'' detector family, GPU/latency limits, etc.); \textbf{(B) an organizational knowledge library} contributes prior tuning outcomes carried across projects (e.g., focal loss unsupported in current YOLO build, class-frequency tables, SOPs); and \textbf{(C) experience from the current evolution} accumulates inside the journal as executed pipelines, mAP metrics, failures (syntax, OOM, NaN), and per-node strength/weakness summaries. (A) and (B) are loaded once per run as the Task / Knowledge / Environmental Context fields and remain fixed thereafter; (C) is dynamic and grows as the search proceeds (\cref{sec:knowledge-update}). This is particularly important for photovoltaic inspection, where prior knowledge remains essential for rigorous and reliable defect analysis.


}

\subsection{Self-Evolving Search Framework}

Building on AIDE~\cite{jiang2025aide} and aira-dojo~\cite{toledo2025ai}, we construct a graph-based search framework with enhanced operators for iterative pipeline optimization. Each expansion creates a new node encapsulating an exploration of the code space (model architecture, training strategy, execution outcome); edges record the evolutionary relationship between successive explorations. This corresponds to the search tree and journal in Fig.~\ref{fig:framework-overview}.

\subsubsection{Components: roles and constraints}
We define the principal components by their role, inputs, outputs, and constraints.

\textbf{Node.} Stores one intra-node code-evolution flow. \emph{Inputs:} parent-node code (or, at root, the SEPDD task description). \emph{Outputs:} candidate code and execution metrics. \emph{Constraint:} a node must pass the per-node debug loop before inter-node expansion, so only validated, fully executed code becomes a search parent.

\textbf{Journal.} Stores all nodes and tree relationships---the complete evolution history of a run---with step-indexed lookup, root query, best-node retrieval, and persistence. As an information hub it exposes a bounded global context (the strategy/result summaries of other valid nodes, fed as the \emph{Summaries of Other Solutions} block in \cref{fig:template}), enabling lightweight inter-node communication so later expansions reuse experience from high-performing nodes without multi-parent merge logic.

\textbf{Operators} are reusable LLM-powered components driving intra-node code evolution. The \emph{Idea Generator} reads parent-node code, execution output, and task requirements (optionally consulting journal summaries) to emit prioritized improvement suggestions (prompt template in \cref{fig:template}). The \emph{Code Creator} translates task description, data description, constraints, and suggestions into a complete executable Python pipeline (initial-generation or improvement mode). The \emph{Analyzer} performs static and dynamic analysis, parses validation/execution output, extracts metrics, and emits structured feedback (\texttt{need\_debug}, \texttt{issues}, \texttt{feedback}). The \emph{Code Refiner} applies targeted fixes guided by Analyzer feedback while avoiding repeated unsuccessful repair strategies. The \emph{Code Evaluator} summarizes evolved code; its output is appended to the journal as the per-node summary that the Idea Generator consults. The \emph{Orchestrator} composes the operators and tools into the per-node flow.

\textbf{Validator and Executor.} The Validator performs syntactic and optional static checks. The Executor runs code in an isolated environment with quick-run / full-run modes, automatic GPU selection, OOM detection with retry, and de-duplicated stdout/stderr.

\textbf{Solver.} Maintains the top-$k$ active set, selects the next node by preferring high-performing candidates with the minimum child count, invokes the Orchestrator, appends new children to the journal, and re-selects the active set.

\subsubsection{Search strategy}
SEPDD adopts a top-$k$ search strategy: at each iteration, the parent for expansion is selected from the current top-$k$ best-performing nodes (exploitation); among those, the node with the minimum child count is expanded (exploration). This relates to Monte Carlo Tree Search but is augmented with a mechanism that reduces node isolation~\cite{du2025automlgen}. \emph{Journal access}---the global journal exposes a bounded shared context \revision{(the strategy/result summaries of the currently best-performing nodes) so that patterns that worked elsewhere influence every new expansion beyond the immediate parent, promoting lightweight cross-branch communication without multi-parent code merging: each new node is derived from a single parent, but its Idea Generator has read the distilled experience of other branches, so effective patterns propagate across branches through the summary channel rather than through explicit code consolidation. This stabilizes the search and reduces drift toward degenerate solutions when feedback is noisy.}

\begin{figure}[t]
\centering
\setlength{\fboxsep}{6pt}
\newcommand{\evotab}{\hspace*{2em}}  
\fcolorbox{black25}{gray8}{%
\begin{minipage}{0.92\linewidth}
\ttfamily\scriptsize
\raggedright
\textbf{Task Context}\\
- Task description\\
- Data description\\
- Task-Specific requirements\\
\smallskip
\textbf{Parent Node}\\
- Code\\
- Execution output\\
- Strategies of this code\\
\smallskip
\textbf{Summaries of Other Solutions}\\
- Model architecture and training strategies\\
- Strength analysis\\
- Weakness analysis\\
\smallskip
\textbf{System Rules}\\
e.g., all task requirements must be satisfied, print metrics explicitly to stdout...\\
\smallskip
\textbf{Reasoning Steps and Instructions}\\
e.g., avoid vague suggestions...\\
\smallskip
\textbf{Output Format}
\end{minipage}%
}
\caption{Prompt template for Idea Generator.}
\label{fig:template}
\end{figure}

\subsubsection{Inter-node pipeline}
Each node in the search graph follows the pipeline:
Idea Generator $\rightarrow$ Code Creator $\rightarrow$ $\text{loop}\{\text{Validator} \rightarrow \text{Analyzer} \rightarrow \text{Executor} \rightarrow \text{Analyzer} \rightarrow \text{Code Refiner}\}$.
The Validator first checks syntax and runs the candidate code in lightweight ``debug'' mode; the Executor then performs the full run; performance metrics and execution traces are fed back to the operators to guide node evaluation and future search decisions. Nodes are ranked by performance, promising nodes are expanded, and branches with two consecutive faulty nodes are terminated to avoid wasted computation. \Cref{alg:node-workflow} formalizes this per-node workflow (\cref{fig:framework-overview}). The debug loop ensures that each node produces functional and evaluated code before becoming a parent and records both successful and unsuccessful attempts to guide subsequent corrections.

\subsubsection{Knowledge update during a run}
\label{sec:knowledge-update}
Updates to (C) (\cref{sec:domain-knowledge}) happen at three points within one run. \emph{(i)~Static load:} (A) and (B) are loaded once per run and remain fixed thereafter. \emph{(ii)~Per-node update:} each expansion runs the full per-node workflow (\cref{alg:node-workflow}); the Idea Generator reads the static context plus \emph{Summaries of Other Solutions} (model architecture, training strategies, strength/weakness analyses of other valid nodes; see \cref{fig:template}), the operators run accordingly, and the journal appends the node, its metrics, and---if valid---its summary, so (C) is strictly enlarged. \revision{\emph{(iii)~Cross-branch update:} the bounded set of summaries is refreshed as new nodes enter the top performers, so the distilled experience of the strongest branches---their effective patterns and failure modes---feeds every subsequent expansion; this summary channel is the consolidation mechanism for (C).} The journal is persisted across long runs, so an operator may pause, edit the static expert/library text (e.g., add a new defect class such as \texttt{blackEdging}), and resume; prior nodes, metrics, and summaries are kept, and only future operator prompts see the revised static context. In our PV runs (\cref{sec:code-insights}), static (A+B) state long-tailed YOLO labels and framework constraints, while (C) later records, e.g., that inverse-square-root oversampling and \texttt{mosaic=0.7} improved tail recall and that focal loss caused training failure---these (C) entries cause subsequent branches to drop focal loss and avoid overly strong mosaic/mixup.

\revision{
\subsection{Operational implementation}
\label{sec:operational-impl}
SEPDD is an offline evolution and training layer, not the runtime inference service. Each successful cycle yields (a) a complete executable training/evaluation pipeline (Python code for data loading, training, metrics) and (b) a trained detector artifact (weights/checkpoint) from the best validated node, both versioned and traceable to the corresponding node and holdout metrics. A separate deployment layer loads the exported model and exposes it as the detector (inference API or embedded service on the EL inspection system), handling preprocessing, batching, thresholds, NMS, logging, and integration with the existing MES/O\&M workflow. In steady state the deployment layer serves the current promoted model on incoming EL images; QC images are expert-labeled and used to detect drift or taxonomy updates that fire SEPDD triggers; after evolution, the new artifact (and inference script/config if needed) is handed off to the deployment layer after validation, and rollback is supported. This separation keeps adaptation (expensive, periodic, LLM-driven pipeline search) distinct from inference (low-latency, stable production path); reported gains in \cref{tab:public,tab:industrial} reflect the quality of the promoted model, while deployment-throughput numbers (\cref{sec:industrial-results}) reflect the production detector layer using that artifact.
}


\section{Experiments} \label{sec:experiments}

This section evaluates SEPDD on both public and private industrial EL defect detection datasets, addressing the key challenges of \cref{sec:problem} (long-tailed labels, distribution shifts, emerging categories). We consider both full-label and reduced-label regimes, and compare SEPDD against AIDE and expert-based baselines. The experiments evaluate the offline-evolved pipeline (\cref{sec:operational-impl}) using held-out, fully labeled test sets; SEPDD reduces repeated manual pipeline tuning but does not remove the need for labels to score and promote candidate models, inheriting the supervised evaluation practice used in conventional industrial PV deployments. \footnote{\revision{The code is available at \url{https://github.com/ecoflow-ai/pv-auto-detection}.}}

\subsection{Experimental Setup}

\subsubsection{Datasets and splits}

\begin{figure}[t]
    \centering
    \begin{subfigure}[t]{0.93\linewidth}
        \centering
        \includegraphics[width=\linewidth]{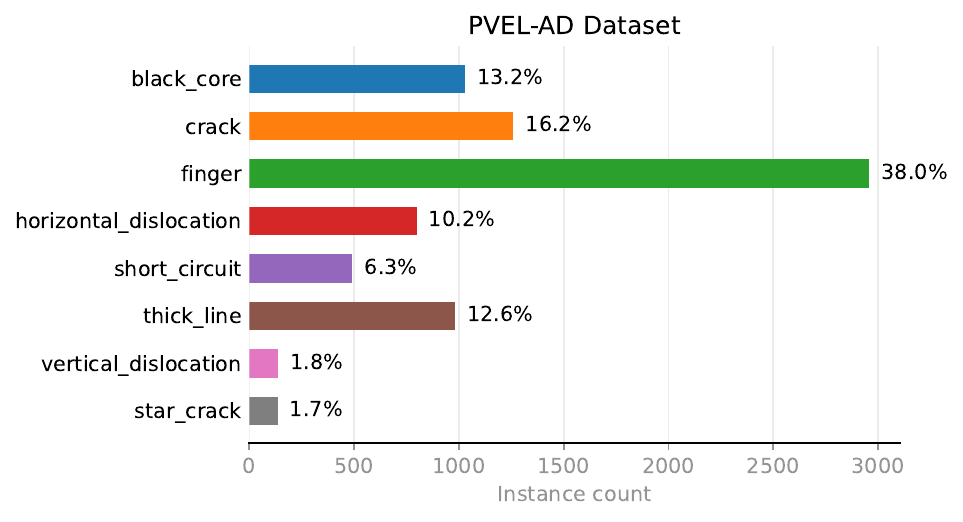}
    \end{subfigure}%

    \begin{subfigure}[t]{0.93\linewidth}
        \centering
        \includegraphics[width=\linewidth]{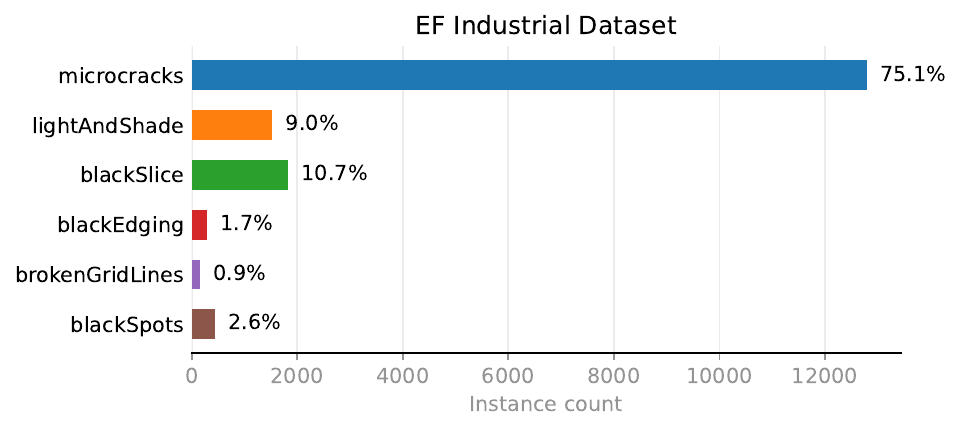}
    \end{subfigure}
    \caption{Defect distribution. \rev{For {EF-Industrial}, train and test splits share a similar long-tailed shape under stratified splitting.}}
    \label{fig:dataset_distribution}
\end{figure}

We use two EL defect detection benchmarks: \revision{the public {\tt PVEL-AD}
\cite{su2022pvel} (4,050 training / 450 test/validation images) and the private {\tt EF-Industrial} (814 training / 90 test/validation images), both in YOLO format with multi-class annotations;} label distributions are in \cref{fig:dataset_distribution}. Compared with {\tt PVEL-AD}, {\tt EF-Industrial} exhibits more heterogeneous panel formats, lower
image resolution, and a substantially more severe long-tailed class distribution, making it a more challenging testbed for robustness 
in industry. 

\emph{EF-Industrial dataset.} Images are collected in a controlled dark-room EL inspection station with a fixed workbench and overhead camera; ambient light is suppressed by design, so the dominant challenges are defect morphology, low contrast, class imbalance, and rare defects rather than unconstrained capture. Annotation proceeds in two stages---experts first define the operational defect taxonomy, then annotators label bounding boxes and class IDs in YOLO format following the protocol used internally for production quality control. The 90-image test set is more diverse than its raw count suggests: each EL image contains multiple defect instances and the corpus covers multiple module models with several frames per module to preserve variation in layout, brightness, and defect appearance. The train/test split is stratified so the long-tailed distribution is preserved across both sets (\cref{fig:dataset_distribution}). \revision{This dataset can be made available to researchers upon reasonable request.}

\subsubsection{Baselines}
We compare three baselines.
\rev{AIDE~\cite{jiang2025aide} is run with the same code-generation backbone as SEPDD (Qwen3-coder-plus) on the same datasets and metrics, but without SEPDD's code-stability components (per-node \cref{alg:node-workflow} debug loop, \revision{journal-augmented top-$k$ search}, multi-operator decomposition), isolating the contribution of these mechanisms rather than backbone differences.}
We further consider two expert baselines bracketing human-driven approaches.
Expert-YOLO is a human-expert-tuned detector based on YOLO11~\cite{yolo11_ultralytics} or YOLO12~\cite{tian2025yolo12} (best result retained), using the same detector family as our pipeline but with manual model selection and tuning.
Expert-FRCN adopts the Faster R-CNN--based few-shot detector~\cite{wang2020frustratingly}, a public-domain detector transferred to industrial PV EL, included to highlight the difficulty of cross-domain adaptation.

\subsubsection{Base model and metrics}
Code generation uses Qwen3-coder-plus, as in AIDE, for fair comparison; analysis operators (Idea Generator, Analyzer) use Qwen3.5-plus. Unless otherwise stated, reported results use this Qwen-based setup. We adopt standard object detection metrics: Precision (P), Recall (R), mAP50, and mAP50-95.


\subsection{Quantitative Results and Analysis}


\begin{table}[t]
    \centering
    \caption{{\tt PVEL-AD} dataset. \rev{SEPDD entries are mean$\pm$std over three independent runs; baselines are single-run.} \textbf{Bold}: best per column; \underline{underline}: second.}
    \label{tab:public}
    \begin{tabular}{lrrrr}
        \toprule
        Method       & Precision & Recall & mAP50 & mAP50-95 \\
        \midrule
        AIDE         & 79.9      & 69.4   & 76.6  & 49.4
        \\
        \midrule
        Expert-YOLO  & 79.9      & 80.4   & 86.7  & 55.7     \\
        Expert-FRCN  & 75.0      & 69.7   & 88.7  & \textbf{63.5}     \\
        \midrule
        SEPDD (Qwen)     & \rev{\textbf{87.0$\pm$4.5}} & \rev{\textbf{87.0$\pm$3.8}} & \rev{\textbf{91.2$\pm$0.4}} & \rev{59.7$\pm$2.5} \\
        SEPDD (GPT-5.1)  & \rev{\underline{86.8$\pm$2.9}} & \rev{\underline{84.1$\pm$1.6}} & \rev{\underline{90.6$\pm$0.4}} & \rev{\underline{61.0$\pm$1.8}} \\
        \bottomrule
    \end{tabular}
\end{table}

\begin{table}[t]
    \centering
    \caption{{\tt \revision{EF-Industrial}} dataset. 
    }
    \label{tab:industrial}
    \begin{tabular}{lrrrr}
        \toprule
        Method       & Precision & Recall & mAP50 & mAP50-95 \\
        \midrule
        AIDE         & 43.4      & 49.2   & 44.6  & 27.7
        \\
        \midrule
        Expert-YOLO  & \textbf{58.4}      & 45.9   & 47.0  & 29.8     \\
        Expert-FRCN  & 38.8      & 32.9   & 36.0  & 18.8     \\
        \midrule
        SEPDD (Qwen)    & \rev{49.2$\pm$3.1} & \rev{\underline{50.4$\pm$3.3}} & \rev{\underline{49.5$\pm$0.3}} & \rev{\textbf{31.2$\pm$0.5}} \\
        SEPDD (GPT-5.1) & \rev{\underline{49.6$\pm$8.5}} & \rev{\textbf{52.0$\pm$4.3}} & \rev{\textbf{49.8$\pm$1.0}} & \rev{\underline{30.8$\pm$0.4}} \\
        \bottomrule
    \end{tabular}
\end{table}

\subsubsection{Public benchmark}
As shown in \cref{tab:public}, the public PVEL-AD dataset is relatively clean and well curated, allowing all methods to achieve competitive performance. AIDE provides a reasonable automated baseline, and Expert-YOLO/Expert-FRCN further improve mAP50.
SEPDD achieves the highest recall and the highest mAP50 among all methods \revision{on this benchmark in our evaluation}.
The substantial recall gain over AIDE suggests that the self-evolving mechanism captures weak, subtle, or partially annotated patterns difficult for a single fixed pipeline to exploit, indicating iterative refinement is beneficial even under idealized conditions.

\subsubsection{EF-Industrial dataset}\label{sec:industrial-results}
Results on the \revision{\tt EF-Industrial} dataset are reported in \cref{tab:industrial}. Compared with the public benchmark, all methods experience a noticeable degradation in absolute performance, consistent with the more challenging industrial conditions (lower resolution, stronger class imbalance, domain shift); Expert-FRCN drops sharply, indicating that a public-domain few-shot detector transfers poorly to private PV EL data without explicit adaptation, consistent with the challenge of distribution shift discussed in \cref{sec:problem}.
SEPDD achieves the best overall performance on this dataset \revision{among the compared methods}, with the largest improvement over AIDE among all methods; the gain is also larger than on the public benchmark, suggesting that the benefit of self-evolution becomes more pronounced when the task is harder and the data are more constrained, supporting the claim that SEPDD is particularly effective under realistic industrial conditions \revision{within the scope of the evaluated datasets and baselines}.

\rev{The detector and pipeline produced by SEPDD have been deployed in our industrial inspection software, reducing the detection time per image from 60\,s to 2\,s and supporting more than 22{,}000 devices over several months of cumulative production throughput; this 22{,}000-device figure reflects deployment-level operational scale and is reported only as deployment evidence, not as a quantitative claim against baselines.}
\revision{Operationally, redeployment is trigger-driven rather than continuous: evolution cycles run offline (with four parallel search workers) only when a \cref{sec:trigger-indicators} indicator fires, and a new artifact is promoted only after it beats the incumbent on the fixed holdout, after which the deployment layer hot-swaps the model (weights plus inference config) with rollback to the previous artifact; in steady state no LLM calls occur on the inference path. In production the deployed detector is monitored through QC-stream sampling (per-class precision/recall on expert-labeled batches), drift statistics on embedding distributions, and inference latency/throughput per station (the 2\,s/image figure corresponds to CPU-batch inference at station rate; the 60\,s figure is the prior manual-inspection baseline). Evolution cycles to date have been triggered mainly by label evolution (new defect categories entering the taxonomy) and periodic refresh, at a cadence of roughly one cycle per one to two months; the remaining operational effort is expert labeling of the QC stream, which the trigger design minimizes by activating statistical checks only on the sampled subset.}

\begin{figure*}[t!]
    \centering
    \begin{subfigure}{0.13\linewidth}
        \centering
        \includegraphics[width=\linewidth]{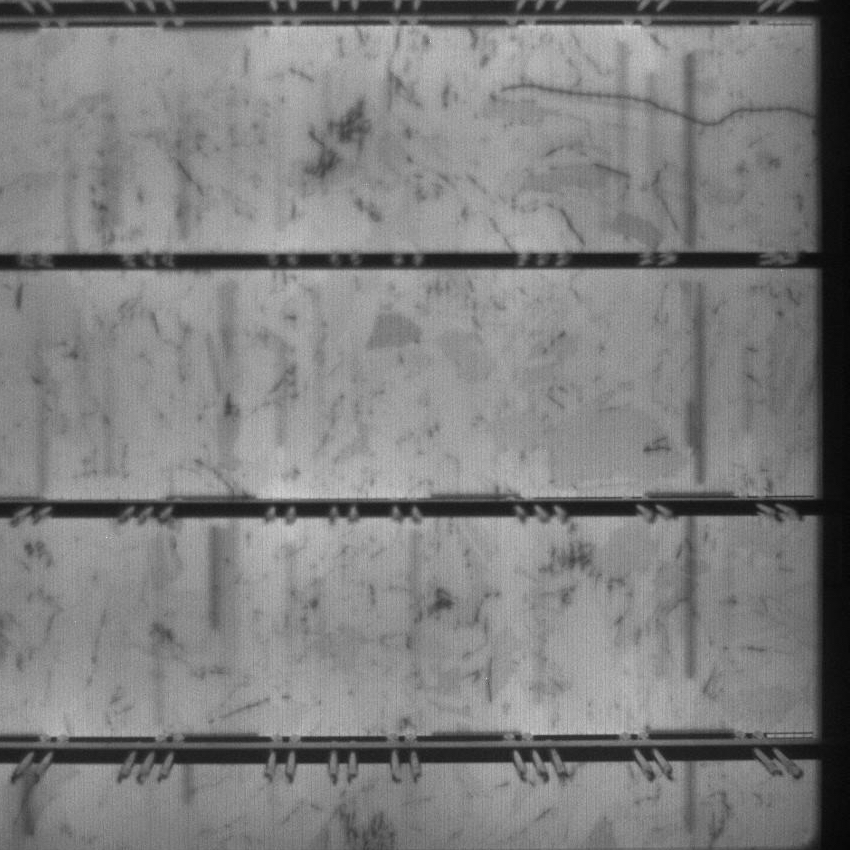}
    \end{subfigure}
    \begin{subfigure}{0.13\linewidth}
        \centering
        \includegraphics[width=\linewidth]{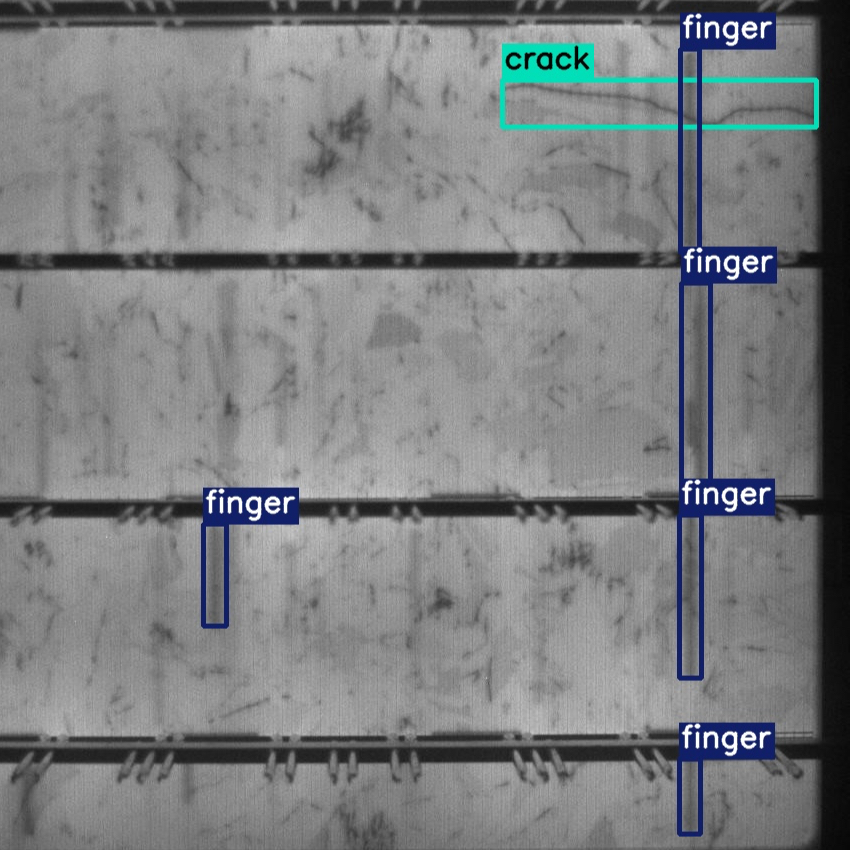}
    \end{subfigure}
    \begin{subfigure}{0.13\linewidth}
        \centering
        \includegraphics[width=\linewidth]{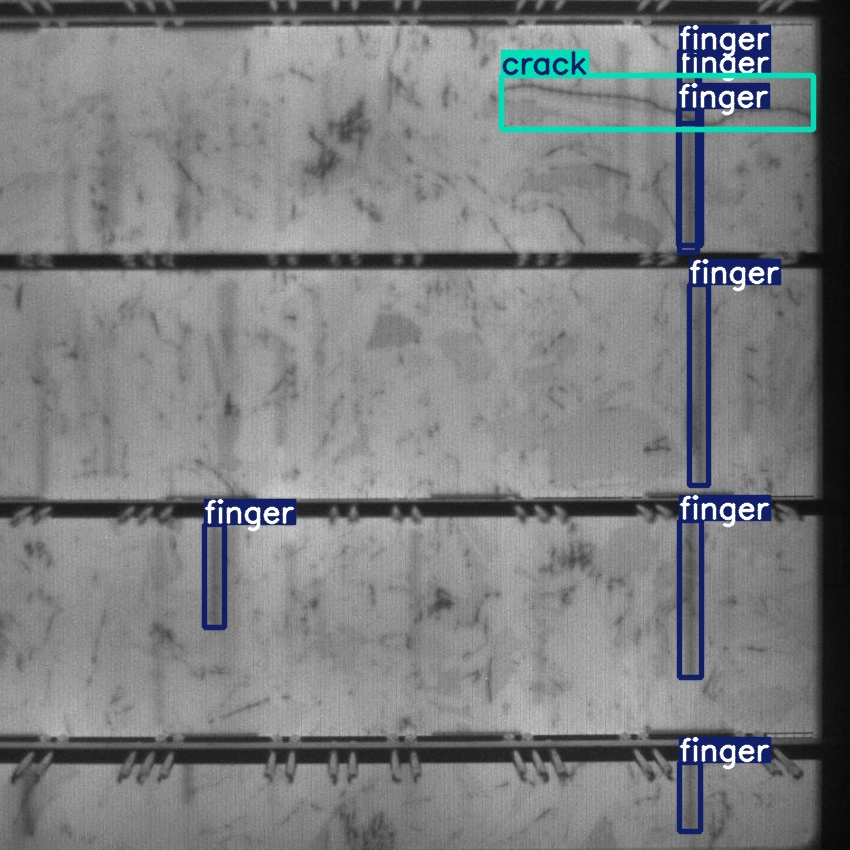}
    \end{subfigure}
    \begin{subfigure}{0.13\linewidth}
        \centering
        \includegraphics[width=\linewidth]{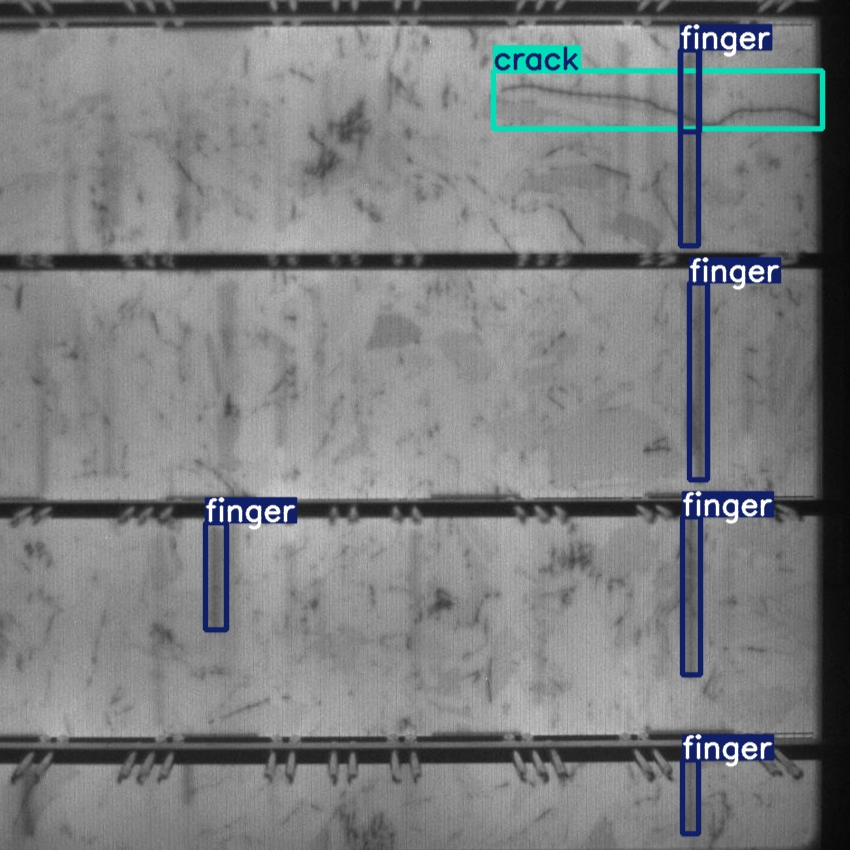}
    \end{subfigure}
    \begin{subfigure}{0.13\linewidth}
        \centering
        \includegraphics[width=\linewidth]{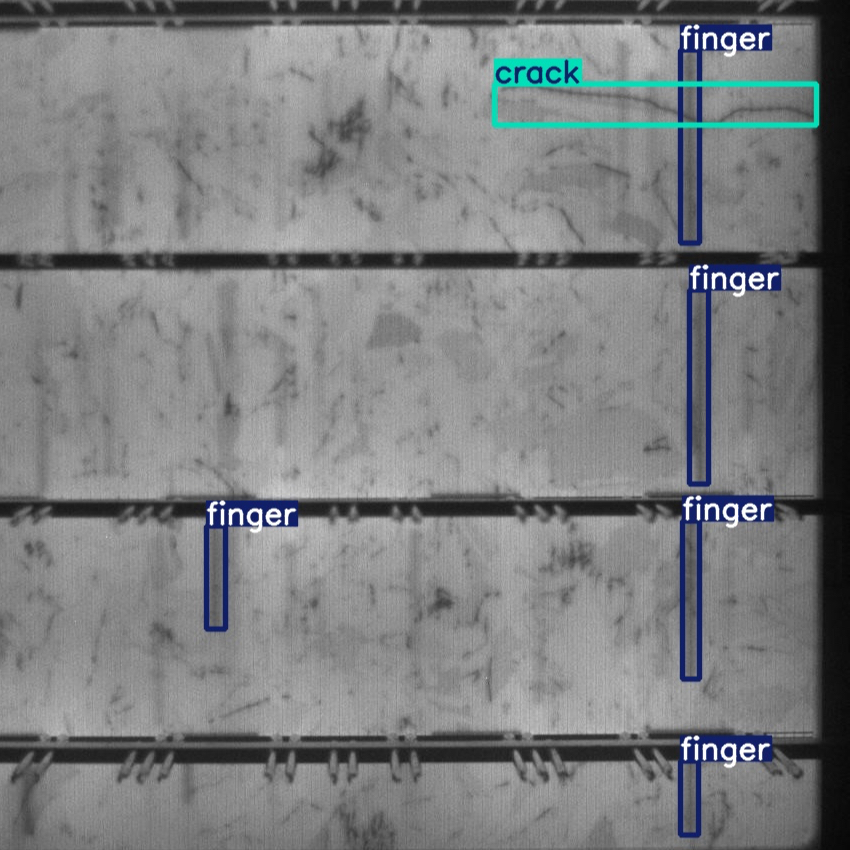}
    \end{subfigure}\vspace{1pt}
    
    \begin{subfigure}{0.13\linewidth}
        \centering
        \includegraphics[width=\linewidth]{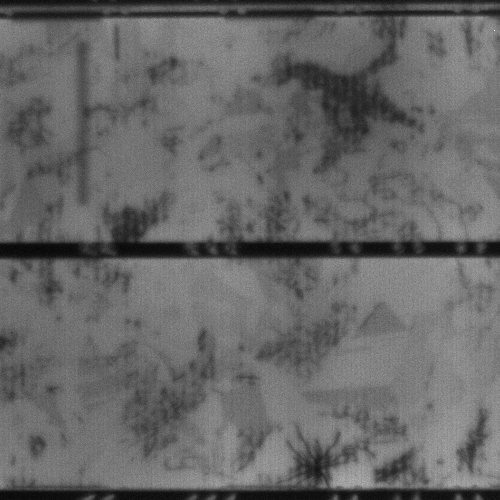}
    \end{subfigure}
    \begin{subfigure}{0.13\linewidth}
        \centering
        \includegraphics[width=\linewidth]{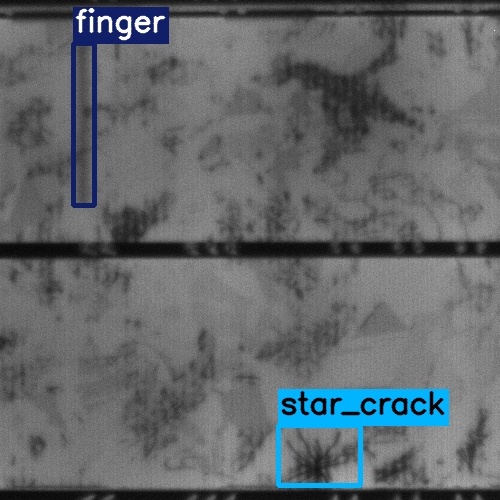}
    \end{subfigure}
    \begin{subfigure}{0.13\linewidth}
        \centering
        \includegraphics[width=\linewidth]{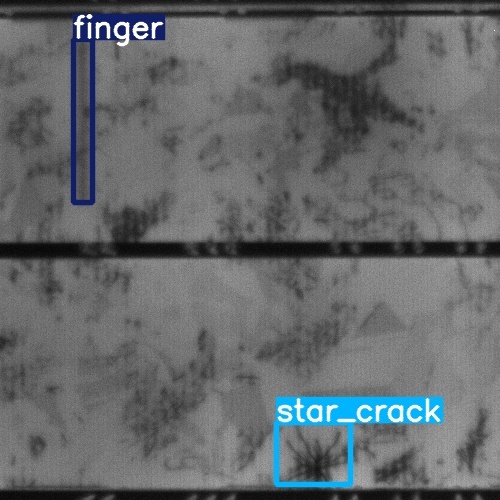}
    \end{subfigure}
    \begin{subfigure}{0.13\linewidth}
        \centering
        \includegraphics[width=\linewidth]{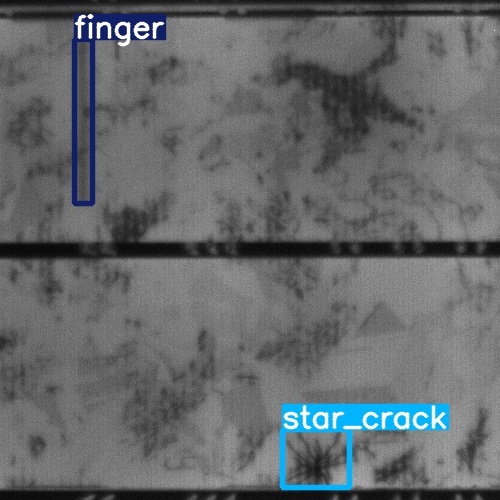}
    \end{subfigure}
    \begin{subfigure}{0.13\linewidth}
        \centering
        \includegraphics[width=\linewidth]{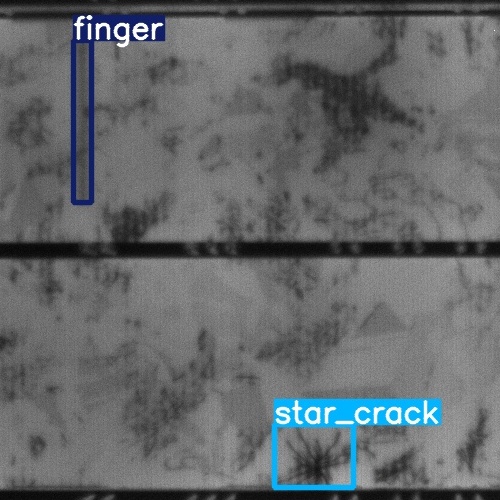}
    \end{subfigure}\vspace{1pt}
    
    
    \begin{subfigure}{0.13\linewidth}
        \centering
        \includegraphics[width=\linewidth]{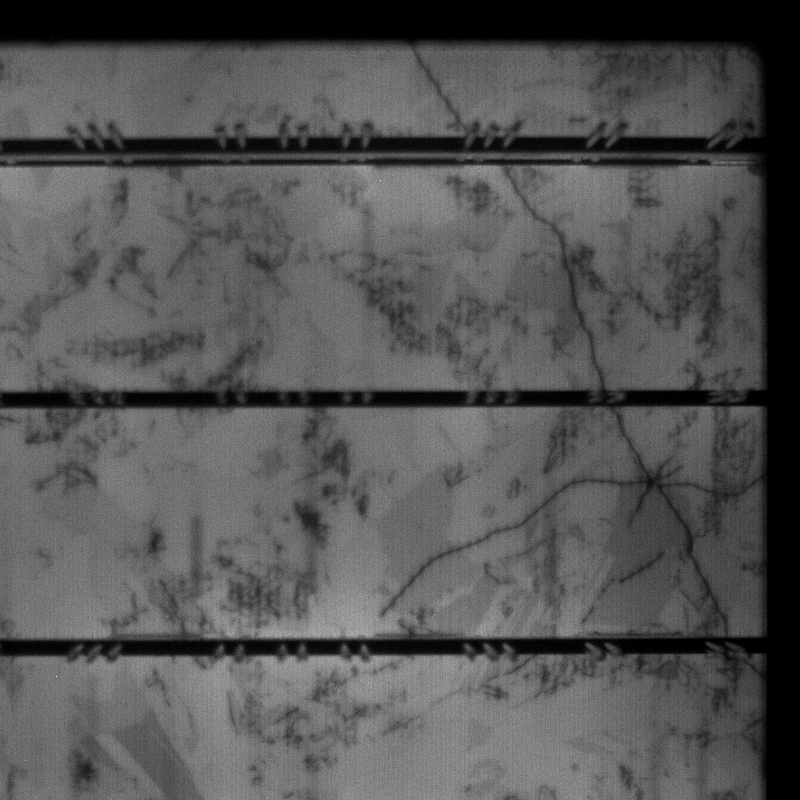}
    \end{subfigure}
    \begin{subfigure}{0.13\linewidth}
        \centering
        \includegraphics[width=\linewidth]{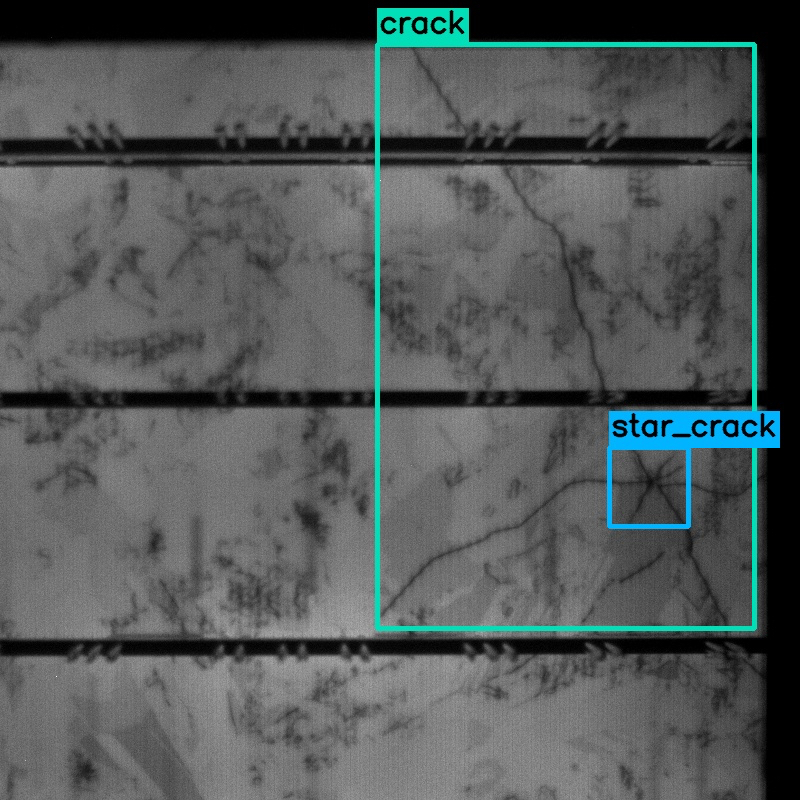}
    \end{subfigure}
    \begin{subfigure}{0.13\linewidth}
        \centering
        \includegraphics[width=\linewidth]{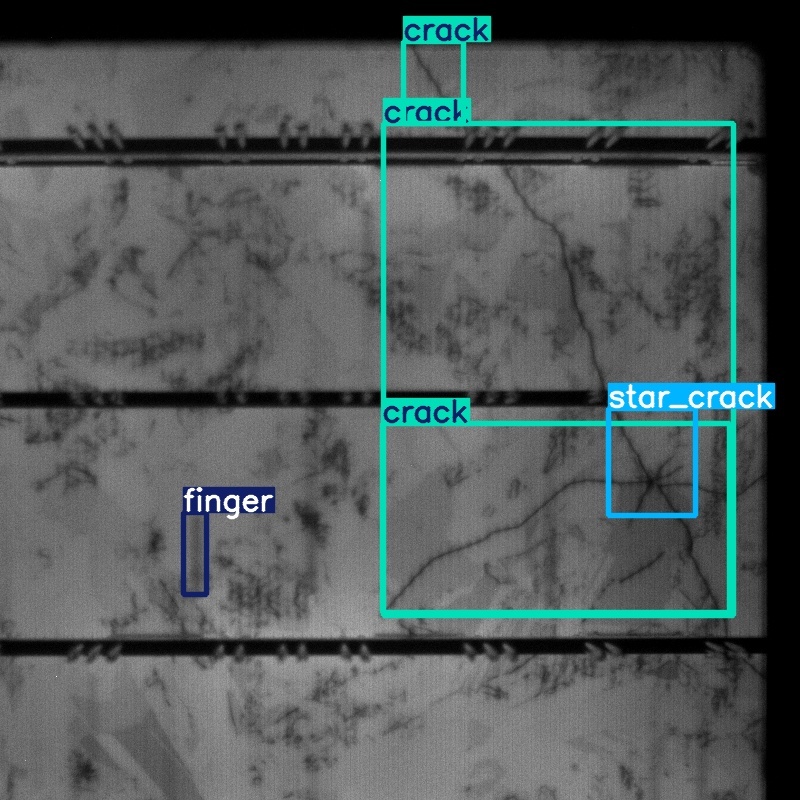}
    \end{subfigure}
    \begin{subfigure}{0.13\linewidth}
        \centering
        \includegraphics[width=\linewidth]{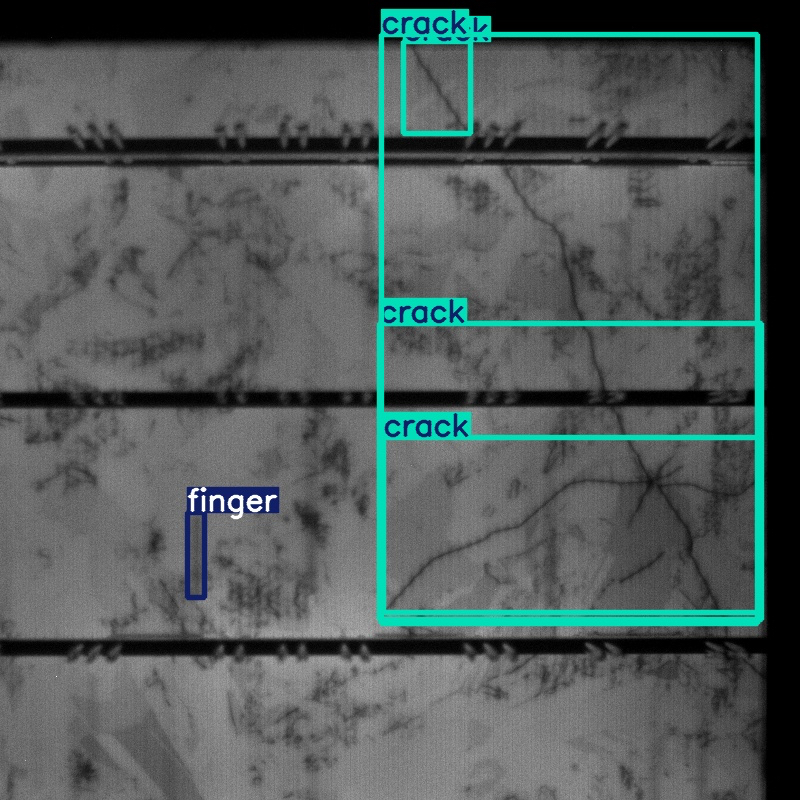}
    \end{subfigure}
    \begin{subfigure}{0.13\linewidth}
        \centering
        \includegraphics[width=\linewidth]{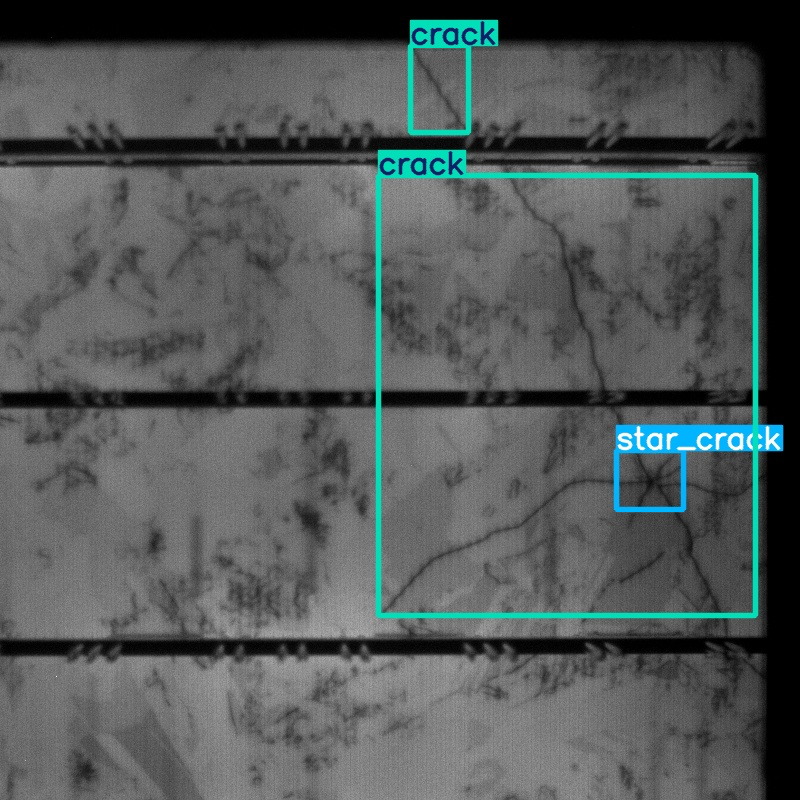}
    \end{subfigure}\vspace{1pt}
    
    \begin{subfigure}{0.13\linewidth}
        \centering
        \includegraphics[width=\linewidth]{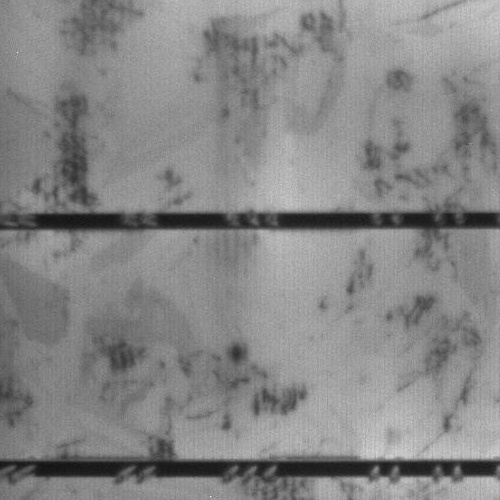}
    \end{subfigure}
    \begin{subfigure}{0.13\linewidth}
        \centering
        \includegraphics[width=\linewidth]{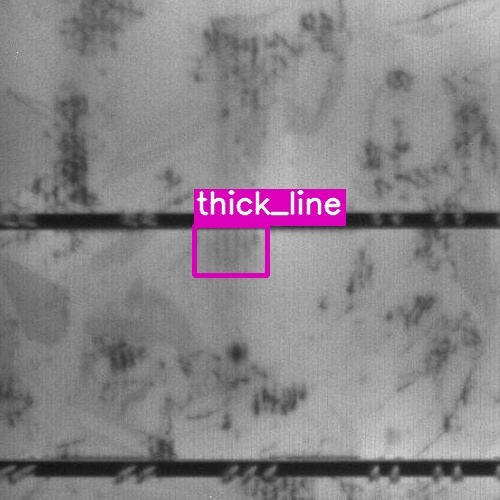}
    \end{subfigure}
    \begin{subfigure}{0.13\linewidth}
        \centering
        \includegraphics[width=\linewidth]{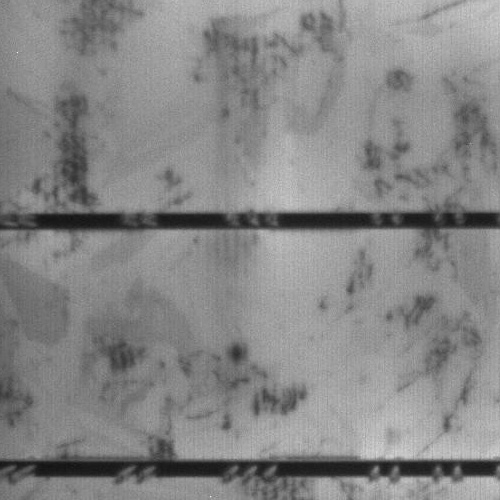}
    \end{subfigure}
    \begin{subfigure}{0.13\linewidth}
        \centering
        \includegraphics[width=\linewidth]{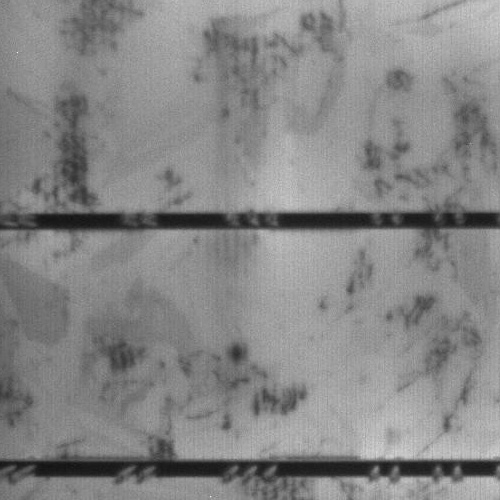}
    \end{subfigure}
    \begin{subfigure}{0.13\linewidth}
        \centering
        \includegraphics[width=\linewidth]{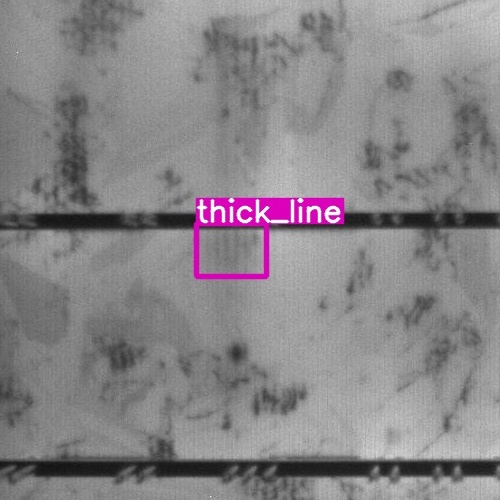}
    \end{subfigure}\vspace{1pt}
    
    \begin{subfigure}{0.13\linewidth}
        \centering
        \includegraphics[width=\linewidth]{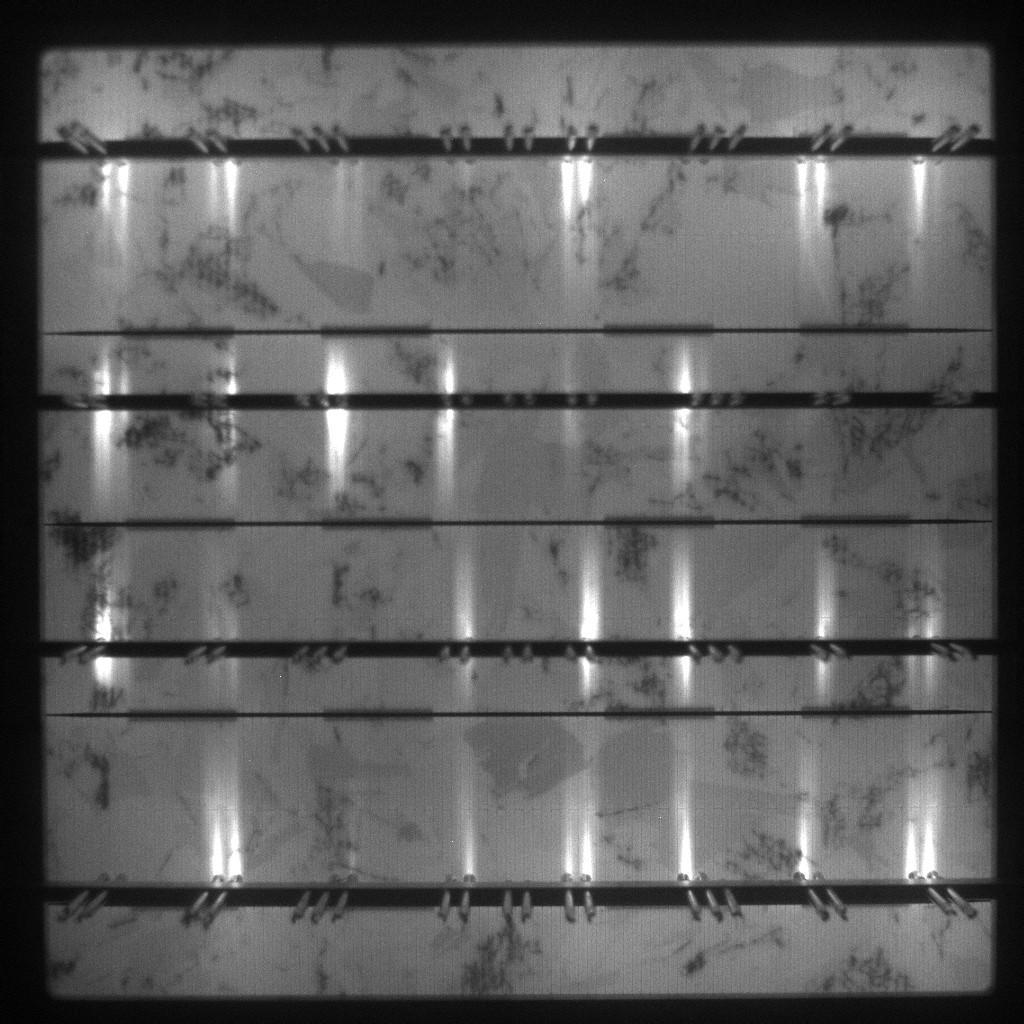}
    \end{subfigure}
    \begin{subfigure}{0.13\linewidth}
        \centering
        \includegraphics[width=\linewidth]{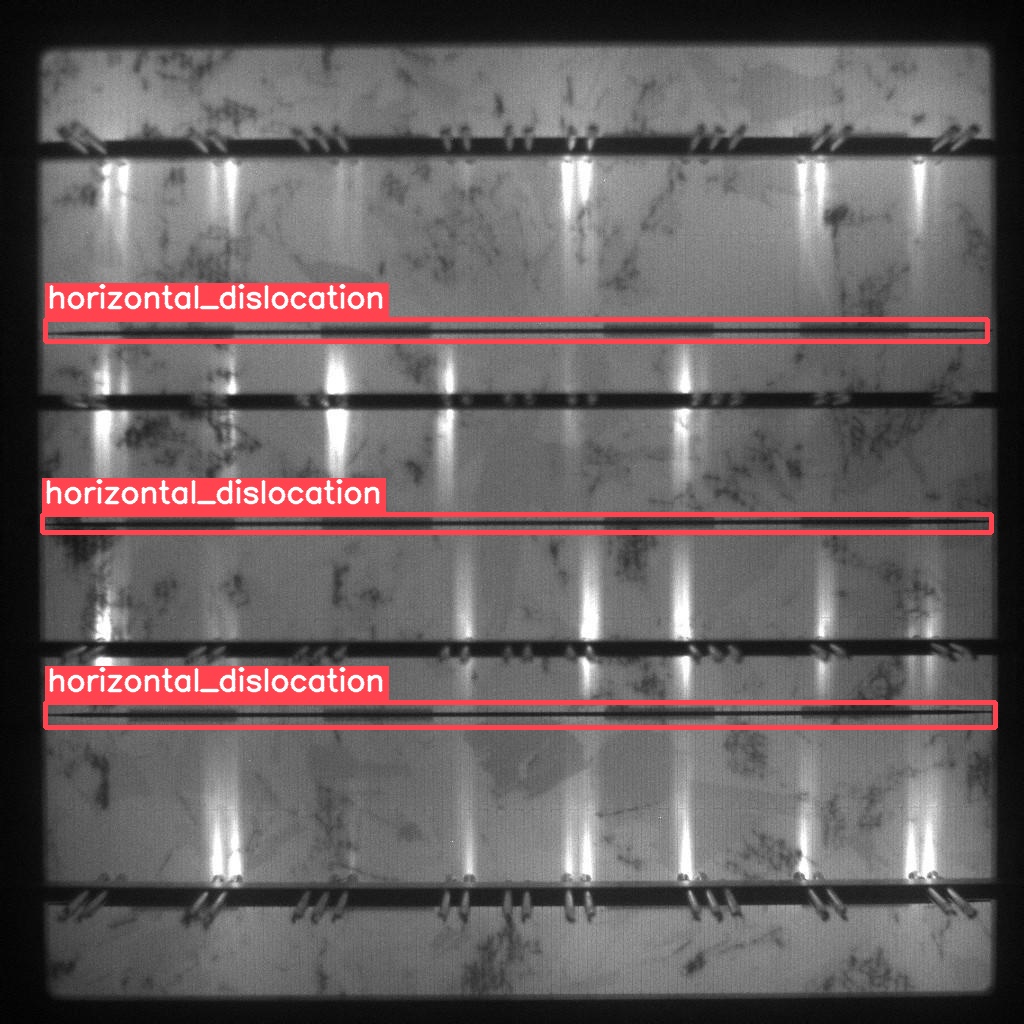}
    \end{subfigure}
    \begin{subfigure}{0.13\linewidth}
        \centering
        \includegraphics[width=\linewidth]{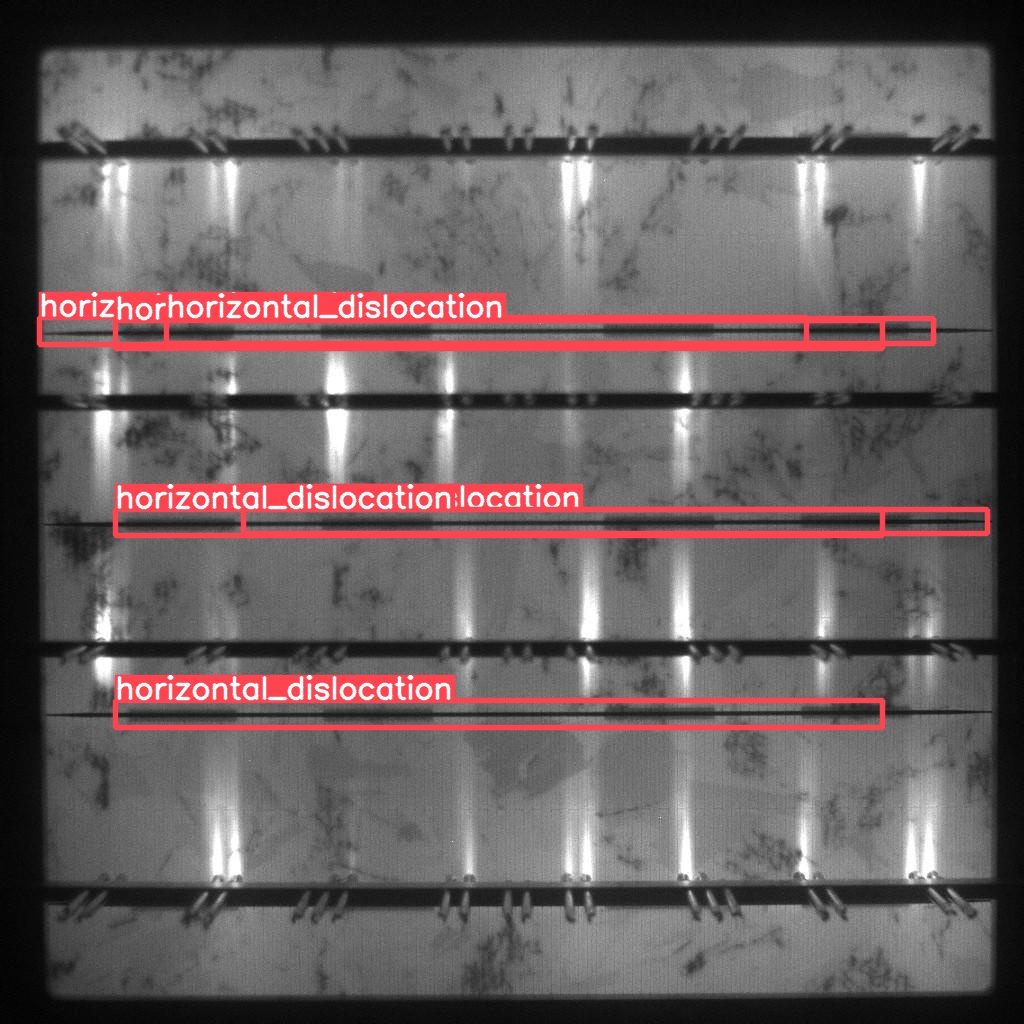}
    \end{subfigure}
    \begin{subfigure}{0.13\linewidth}
        \centering
        \includegraphics[width=\linewidth]{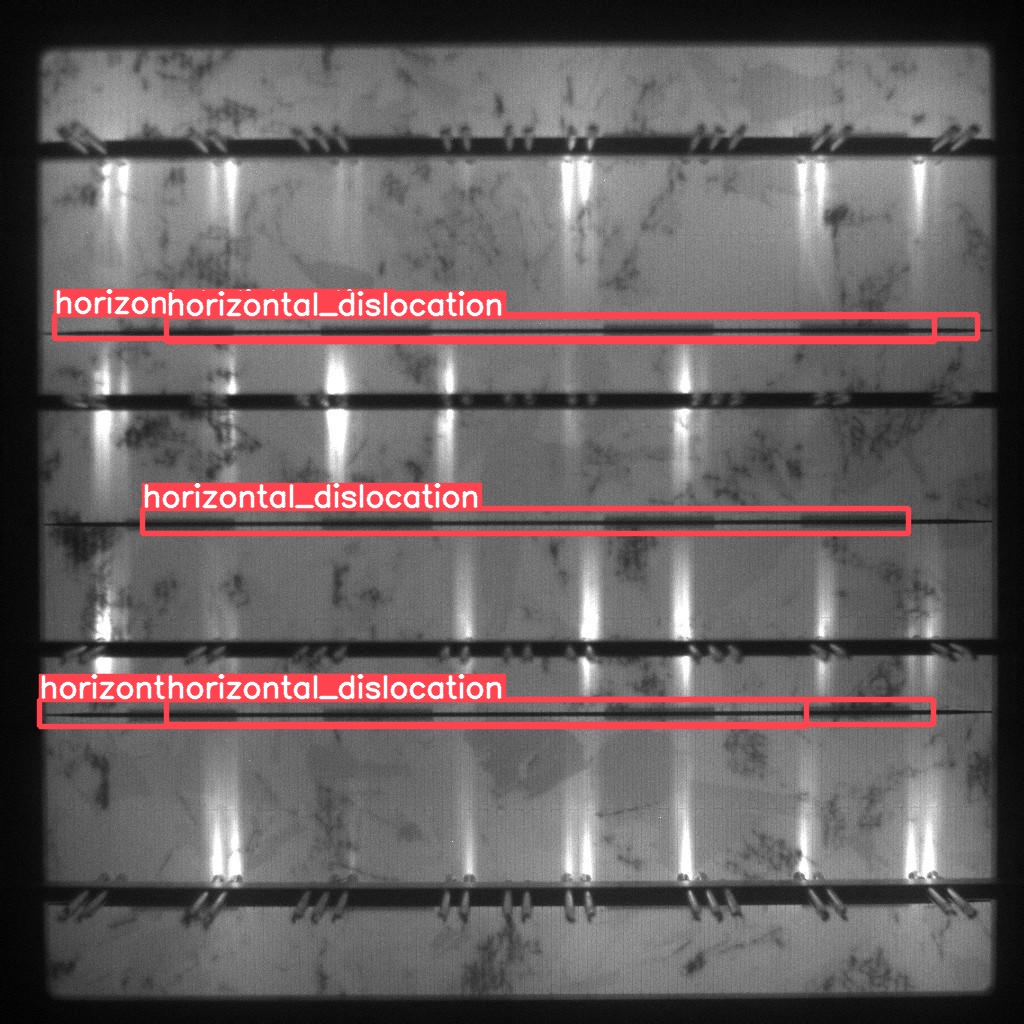}
    \end{subfigure}
    \begin{subfigure}{0.13\linewidth}
        \centering
        \includegraphics[width=\linewidth]{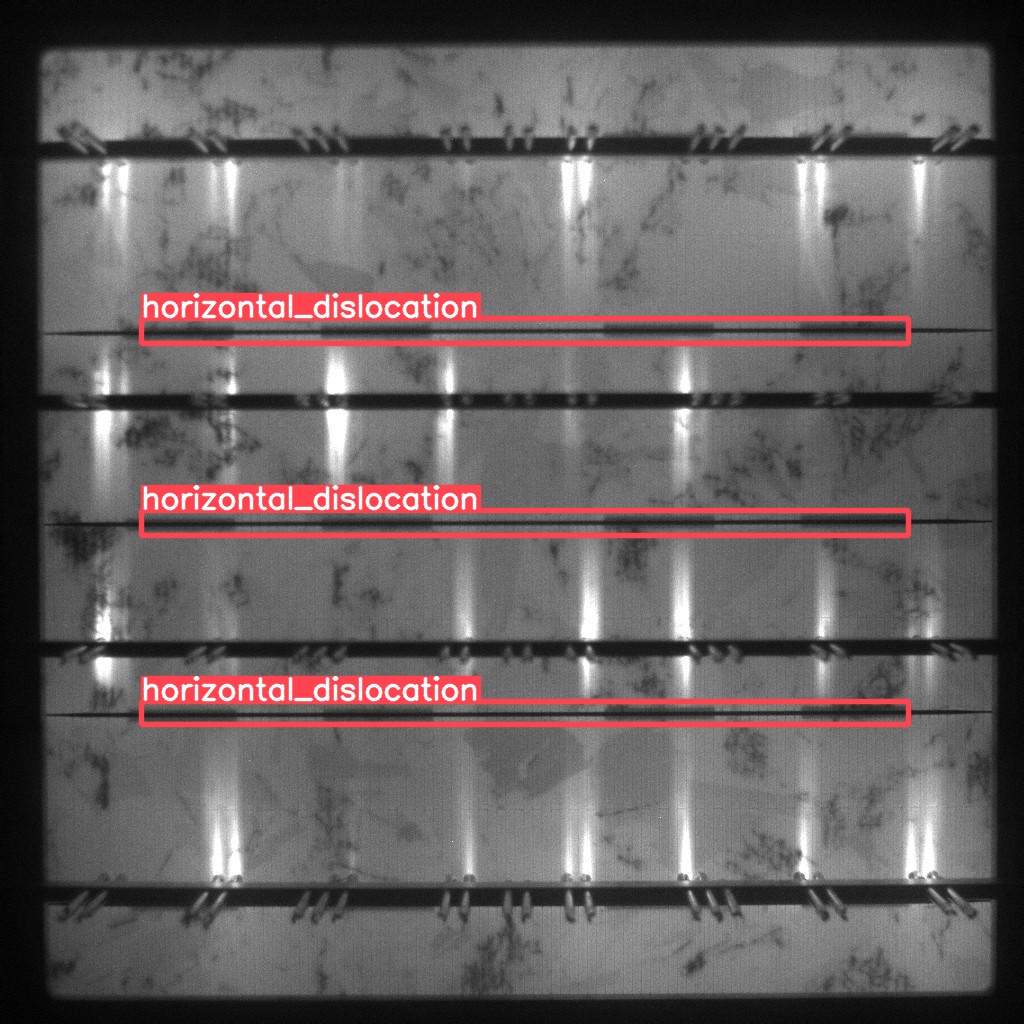}
    \end{subfigure}\vspace{1pt}
    \caption{Detection comparison across public defect categories. Each row: defect type; each column: Original, Ground truth, AIDE, Human Expert (Expert-YOLO), Self-Evolving (SEPDD).}
    \label{fig:8x5_public}
\end{figure*}

\subsubsection{Visualization and Observation}
\begin{figure*}[t]
    \centering
    \newcommand{\efrowlabel}[1]{\scriptsize\textbf{#1}}
    \newcommand{\efimgheight}{5em}
    \setlength{\tabcolsep}{2pt}
    \renewcommand{\arraystretch}{0}
    \begin{tabular}{@{}>{\centering\arraybackslash}m{0.08\linewidth}@{\hspace{4pt}}ccc@{}}
        \efrowlabel{Ground truth} &
            \includegraphics[height=\efimgheight]{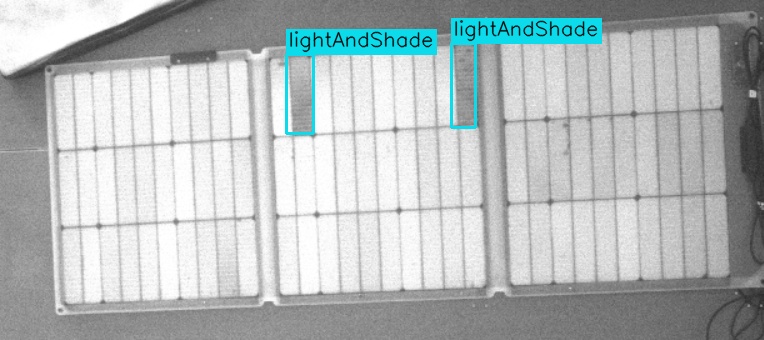} &
            \includegraphics[height=\efimgheight]{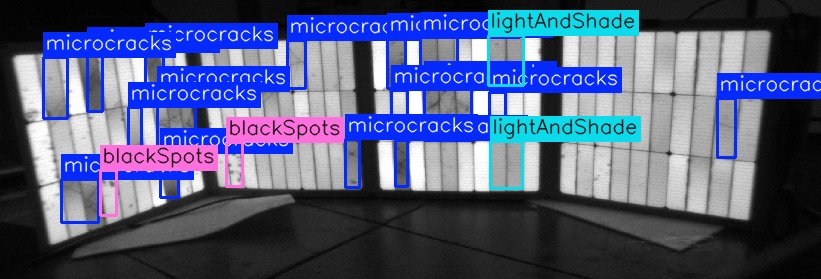} &
            \includegraphics[height=\efimgheight]{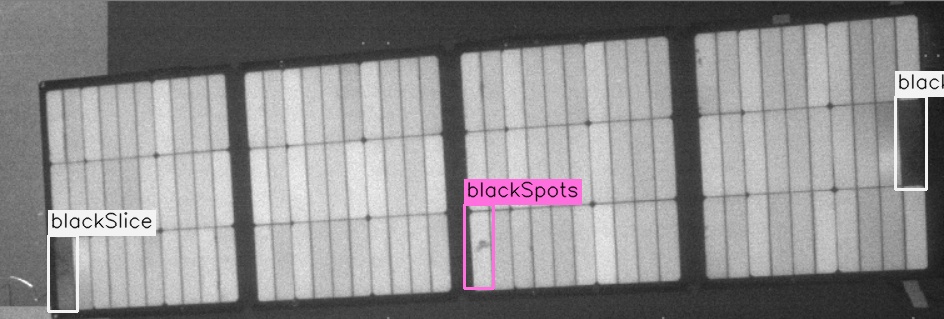} \\
        \efrowlabel{AIDE} &
            \includegraphics[height=\efimgheight]{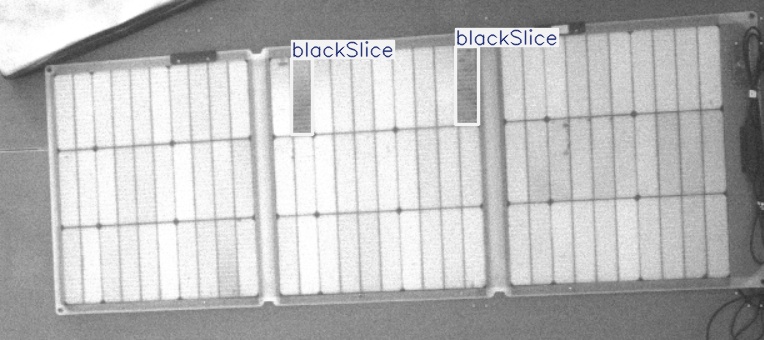} &
            \includegraphics[height=\efimgheight]{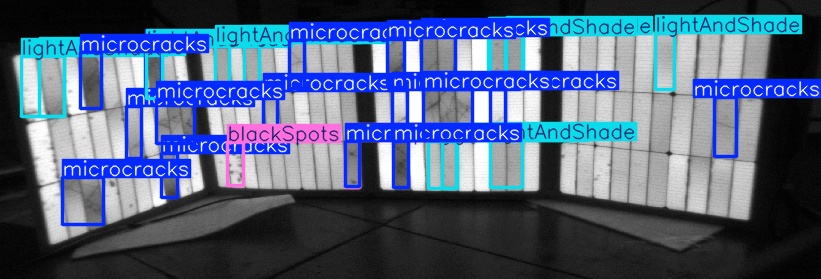} &
            \includegraphics[height=\efimgheight]{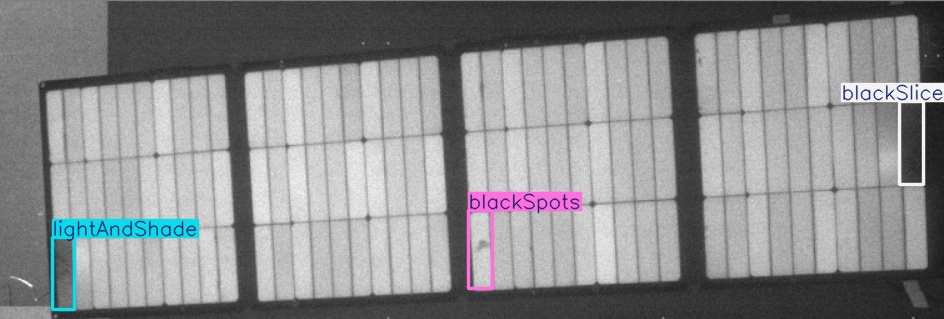} \\
        \efrowlabel{Human Expert(Expert-YOLO)} &
            \includegraphics[height=\efimgheight]{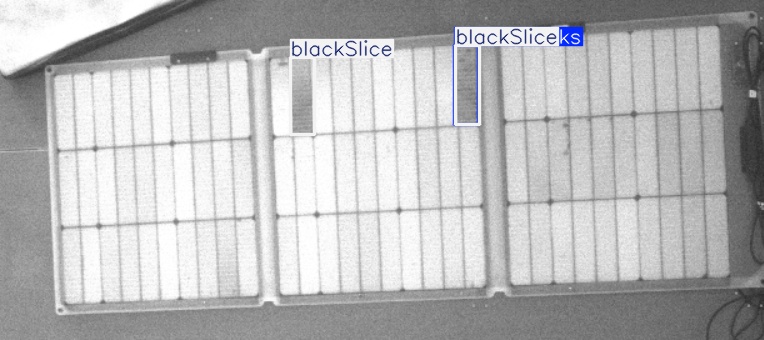} &
            \includegraphics[height=\efimgheight]{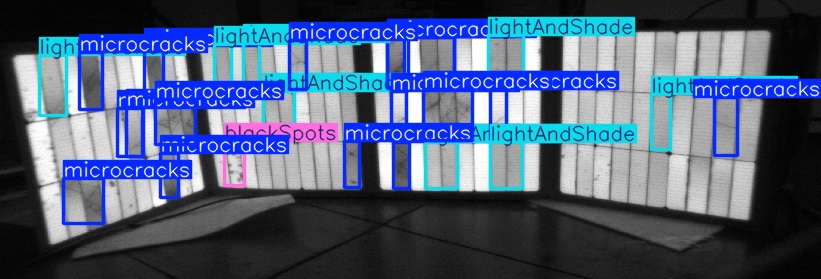} &
            \includegraphics[height=\efimgheight]{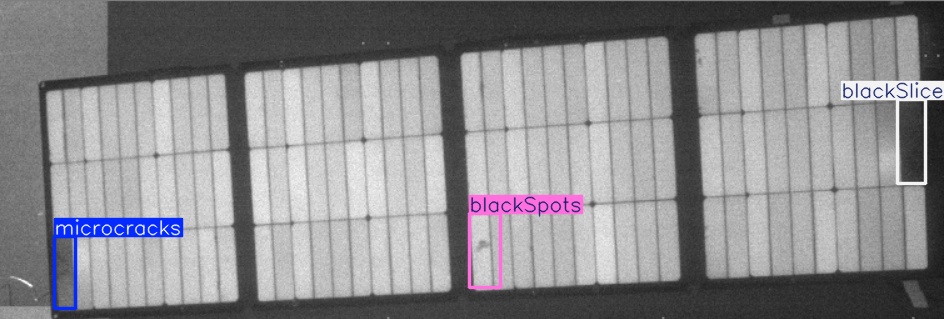} \\
        \efrowlabel{SEPDD} &
            \includegraphics[height=\efimgheight]{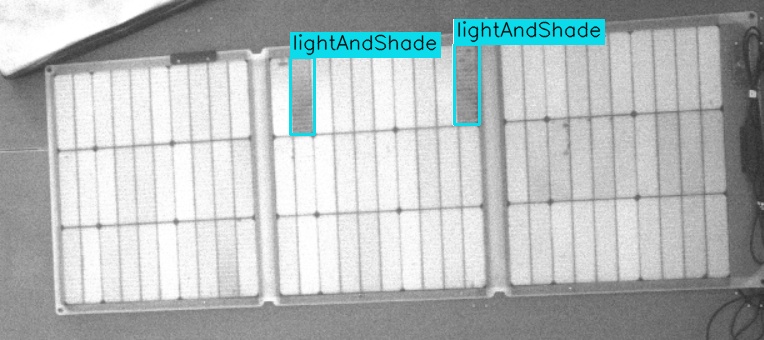} &
            \includegraphics[height=\efimgheight]{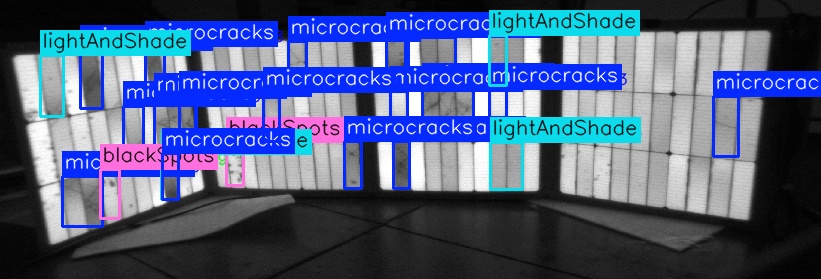} &
            \includegraphics[height=\efimgheight]{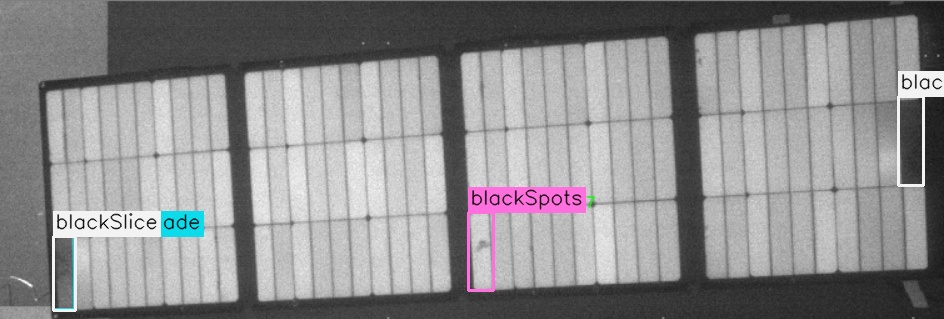} \\
    \end{tabular}
    \caption{Detection comparison on representative EF industrial scenes (columns: dominant defect type per scene---light/shade, black spot, black slice; each scene contains multiple defect instances of varying types). 
    }
    \label{fig:5x5_industrial}
\end{figure*}

Visual comparisons complement the quantitative results. \Cref{fig:8x5_public} compares detection outputs on the public benchmark across defect categories (rows: defect type; columns: original, ground truth, AIDE, Expert-YOLO, SEPDD); \Cref{fig:5x5_industrial} presents representative industrial examples.
On the public dataset, all methods generally identify the major defects, but SEPDD more often recovers subtle or low-contrast defects that AIDE and---in some cases---the expert-tuned models miss (e.g., thick line in \cref{fig:8x5_public}).
On the industrial dataset, the advantage is more pronounced: under lower resolution and stronger long-tailed imbalance, SEPDD maintains better coverage of weak or ambiguous defects (e.g., light/shade vs.\ black slice in \cref{fig:5x5_industrial}), consistent with the gains in \cref{tab:industrial}.
These qualitative observations align with the central goal of SEPDD: improving defect coverage and robustness under the same challenges emphasized throughout this paper, namely complex morphology, domain shift, and evolving defect patterns.

\subsubsection{Repeatability of LLM-based search}
mAP50 standard deviations stay below 1.0 across all configurations in \cref{tab:public,tab:industrial} and mAP50-95 within roughly 0.4--2.5 percentage points; precision/recall fluctuate more, but the detector-level metrics that drive the comparison are stable, and evolved pipelines converge to similar design choices across runs (consistent oversampling and augmentation, similar backbone--learning-rate coupling). Two factors contribute: (i) a clear holdout signal gives the search a reliable gradient; and (ii) the per-node Validator/Analyzer/Executor/Code Refiner loop and the \revision{journal's summary channel} filter unstable candidates before they affect the search.

\subsubsection{Tail-class scarcity proxy (one-label-removed)}

\begin{figure}[t]
    \centering
    \begin{subfigure}[t]{0.93\linewidth}
        \centering
        \includegraphics[width=\linewidth]{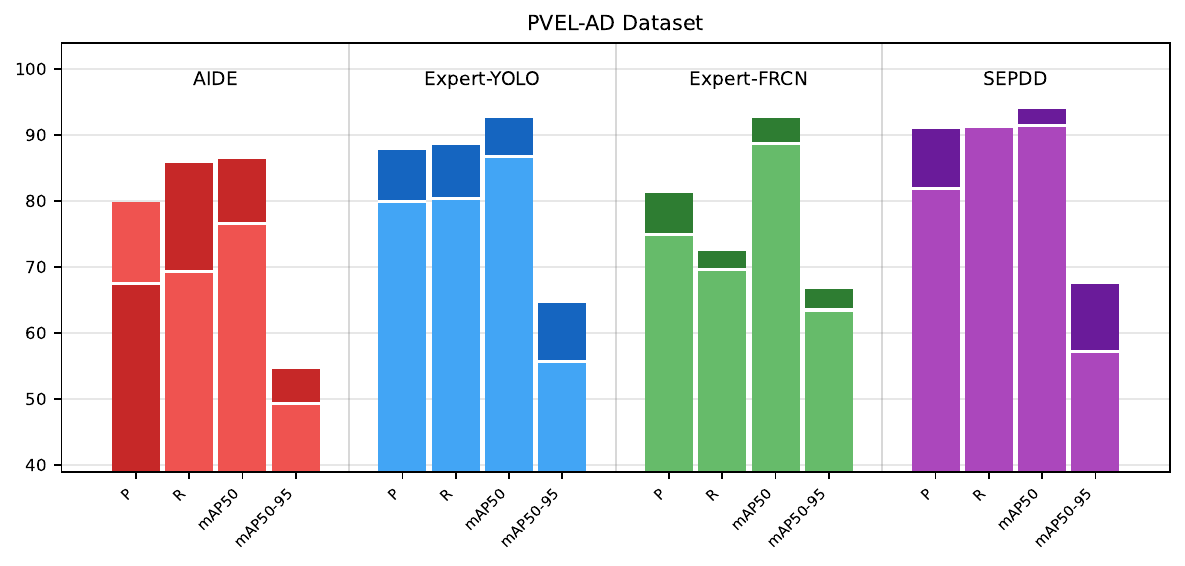}
    \end{subfigure}%
    
    \begin{subfigure}[t]{0.93\linewidth}
        \centering
        \includegraphics[width=\linewidth]{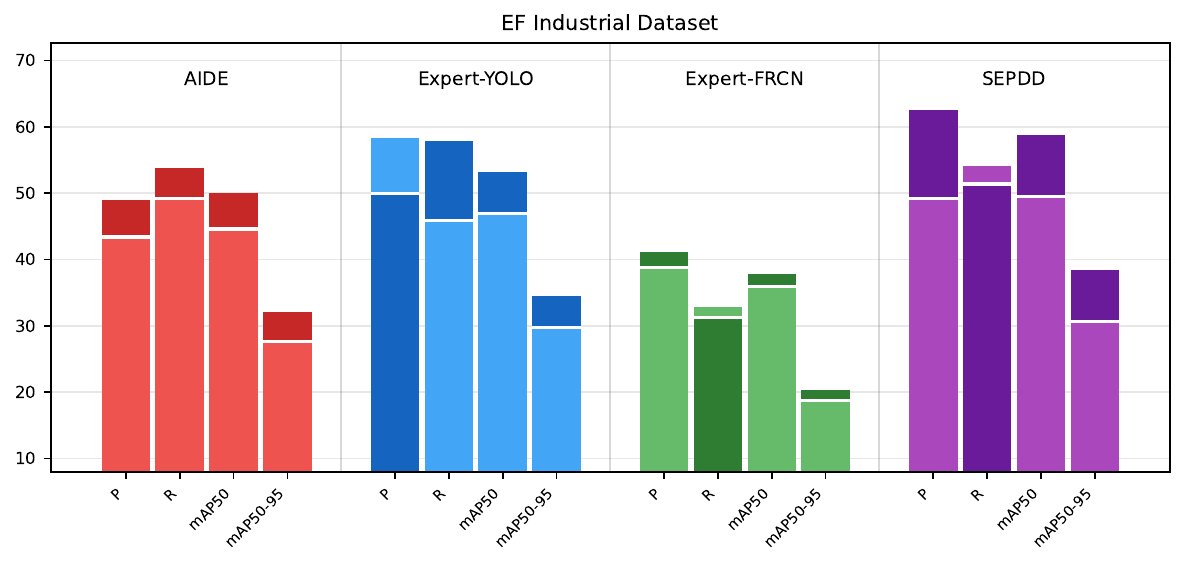}
    \end{subfigure}
    \caption{\rev{Tail-class scarcity proxy: full label (lighter, the state after the rare class is included) vs.\ one label removed (darker, the prior state). \texttt{star\_crack} on PVEL-AD; \texttt{blackEdging} on EF-Industrial.}}
    \label{fig:one-label-regime}
\end{figure}

In real-world industrial settings, we often encounter emerging defect categories of small instance counts (Challenge 3). \rev{To stress-test whether automated pipeline evolution remains effective when supervision for a reliability-critical tail class is scarce---an operational pattern that often precedes formal registration of a new defect category---we simulate the prior state via a one-label-removed proxy:} we remove one low-frequency label from the training set (\texttt{star\_crack} on PVEL-AD, \texttt{blackEdging} on EF-Industrial) while keeping the test taxonomy fixed and re-run all methods.
The results are shown in \cref{fig:one-label-regime}, where darker bars correspond to the one-label-removed setting (prior state) and lighter bars to the full-label setting (after the class is added).
\rev{This is a single-step proxy for scarce tail-class supervision, not a multi-stage continual-learning evaluation and not a direct measurement of catastrophic forgetting.}
On both datasets, SEPDD maintains its performance better than the baselines when a rare label is removed and later treated as newly introduced; AIDE and Expert-FRCN degrade more noticeably, especially on the industrial scenario. Because SEPDD automates pipeline search within the self-evolving loop, it can adapt to new categories without sacrificing overall robustness; across both settings (\cref{tab:public,tab:industrial,fig:one-label-regime}), SEPDD consistently matches or outperforms the baselines in mAP50 and mAP50--95.

\subsection{Code Strategy and Pipeline Analysis and Insights}

The gains of SEPDD over AIDE and the expert baselines arise from an autonomous training pipeline (shared by AIDE and SEPDD, already incorporating PV EL domain knowledge) further improved through iterative self-evolution (\cref{sec:methodology}).

\subsubsection{Code-level insights}
\label{sec:code-insights}

The challenges in \cref{sec:problem} motivate the directions in which the code evolves; the insights below are the resulting design choices.

\begin{itemize}
    \item \textit{Long-tailed and limited data.}
The search favors configurations that better represent minority defect categories: \texttt{cls}=1.0 with inverse-square-root-weighted oversampling (focal loss is unsupported in the current framework); stronger augmentation (mosaic=0.7, mixup=0.15, degrees=5, translate=0.10, scale=0.5); AdamW with \texttt{lr0}=8e-4 and cosine decay; a larger backbone (e.g., YOLO11l) only when paired with a matching augmentation/optimization recipe.
    \item  \textit{Complex morphology and distribution shift.}
The search converges to PV-specific, robustness-oriented choices: augmentation strong enough to improve generalization without displacing tiny targets (microcracks, gridline breaks); moderate rotation/translation; an input size of 640; a stronger box loss and label smoothing for bounding-box quality and calibration.
    \item \textit{What did not work.}
The search avoids focal loss (unsupported, training fails), very high input resolution (e.g., 896), excessively strong mosaic/mixup, and a larger backbone without matching learning rate and augmentation.
\end{itemize}

\subsubsection{Self-evolving progress}

When distilling information across the search graph, SEPDD keeps only the \emph{primary edge}---the best parent-child lineage along the current best branch.
\Cref{fig:evolution-tree} shows the evolution tree from one EF-Industrial run: 18 nodes total, with the best mAP50 at Node 10 (0.4954) along the primary edge $0\to1\to6\to10$ (Node 11 is invalid; the branch under Node 2 does not surpass the best branch). The detector and metrics in the main tables correspond to the best node selected from such a run.

This behavior highlights several properties of SEPDD: \revision{the journal and its summary channel reduce search drift; strong gains are achieved with a shallow tree}; domain knowledge enters through updated knowledge inputs, task-specific triggers, and specialized operators; and the design substantially reduces the repeated learn--tweak--try cycle of manual expert tuning.
By contrast, AIDE~\cite{jiang2025aide} may produce a much larger solution tree (often $\geq$50 nodes) with many branches devoted to drafting, debugging, and refining code. SEPDD records only workable nodes and achieves strong results with a shallower tree, consistent with its design objective of stable, efficient self-evolution.

\begin{figure}[t]
\centering
\setlength{\fboxsep}{3pt}
\newcommand{\evotab}{\hspace*{1.6em}}  
\fcolorbox{black25}{gray8}{%
\begin{minipage}{0.95\linewidth}
\ttfamily\tiny
\raggedright
Root 0\\
+-{}- \textbf{1: mAP50-95=0.2764, mAP50=0.4433}\\
|\evotab+-{}- 5: mAP50-95=0.4106, mAP50=0.4106\\
|\evotab{}L-{}- \textbf{6: mAP50-95=0.2970, mAP50=0.4578}\\
|\evotab~\evotab+-{}- 9: mAP50-95=0.3051, mAP50=0.4592\\
|\evotab~\evotab|\evotab+-{}- 13: mAP50-95=0.3045, mAP50=0.4609\\
|\evotab~\evotab|\evotab|\evotab+-{}- 17: mAP50-95=0.2942, mAP50=0.4561\\
|\evotab~\evotab|\evotab|\evotab{}L-{}- 18: mAP50-95=0.2935, mAP50=0.4736\\
|\evotab~\evotab|\evotab{}L-{}- 14: mAP50-95=0.2552, mAP50=0.3919\\
|\evotab~\evotab{}L-{}- \textbf{10: mAP50-95=0.3069, mAP50=0.4954}\\
|\evotab~\evotab~\evotab+-{}- 11: mAP50-95=-{}--, mAP50=-{}-- (invalid)\\
|\evotab~\evotab~\evotab+-{}- 12: mAP50-95=0.2767, mAP50=0.4549\\
|\evotab~\evotab~\evotab+-{}- 15: mAP50-95=0.2777, mAP50=0.4581\\
|\evotab~\evotab~\evotab{}L-{}- 16: mAP50-95=0.2904, mAP50=0.4909\\
L-{}- 2: mAP50-95=0.2718, mAP50=0.4325\\
~\evotab+-{}- 3: mAP50-95=0.2827, mAP50=0.4723\\
~\evotab{}L-{}- 4: mAP50-95=0.2685, mAP50=0.4106\\
~\evotab~\evotab+-{}- 7: mAP50-95=0.2802, mAP50=0.4389\\
~\evotab~\evotab{}L-{}- 8: mAP50-95=0.2433, mAP50=0.4114
\end{minipage}%
}
\caption{Evolution tree from one EF-Industrial run. \textbf{Bold}: best branch $0 \to 1 \to 6 \to 10$.}
\label{fig:evolution-tree}
\end{figure}
\subsection{Additional study}

\subsubsection{Ablation setting}
We examine four variants of SEPDD:
\begin{itemize}
    \item \textit{Greedy (SEPDD (greedy))}: always selects the locally best candidate without exploring alternative branches.
    \item \revision{\textit{Without journal summaries (SEPDD w/o journal)}: the cross-branch communication channel is disabled---the Idea Generator no longer receives the \emph{Summaries of Other Solutions} block---while} the journal as a storage container is retained.
    \item \textit{Both greedy and \revision{no journal summaries} (SEPDD (both))}: combination of the above.
    \item \textit{Without refinement (SEPDD w/o refinement)}: disables the per-node Validator/Analyzer/Executor/Code Refiner loop, isolating the contribution of per-node refinement.
\end{itemize}
\subsubsection{Performance}
The differences for \emph{greedy}, \revision{\emph{w/o journal}}, and \emph{both} are moderate under our limited evolution depth, while the increased token usage in \cref{tab:toks} indicates extra resources spent on refinement.
In contrast, removing per-node refinement is highly destructive: the \emph{w/o refinement} variant produced only \textbf{2~valid (executable, metric-bearing) nodes} across the full search budget on PVEL-AD; the rest contained bugs the debug loop would normally have caught, and the loop's debug traces were therefore not captured. The numbers in \cref{tab:ablation} for \emph{w/o refinement} reflect the best of these two valid nodes; mAP50 drops from 91.5 to 90.1 and the search effectively stalls---\revision{the journal summaries have} almost no valid material to draw from---confirming that refinement is the dominant stabilizer of the per-node search state. We therefore report \emph{w/o refinement} alone rather than in combinations.
\revision{Because the \emph{w/o journal} row sits within run-to-run noise of the full configuration, we additionally re-ran the \emph{full} vs.\ \emph{w/o journal} comparison as a paired study on PVEL-AD: independent runs per configuration with identical LLM backbones and budgets, differing only in whether the summary channel is disabled. The paired results (best 89.7 vs.\ 89.3 mAP50; 2 valid nodes each, out of 18 and 20 nodes) confirm that the summary channel does not measurably improve single-cycle peak accuracy on this benchmark: within one shallow cycle the cross-branch summaries do not raise the search ceiling, and the per-node refinement loop remains the dominant stabilizer. The journal's measurable benefit appears in settings where accumulated experience has material to reuse---deeper search and warm-started multi-round evolution (\cref{sec:multiround}), where the summary channel propagates effective patterns across branches and rounds.}
Triggers determine \emph{when} an evolution cycle starts; the experiments here study search behavior \emph{after} a cycle has been triggered, so a literal trigger-removal reduces to a static pipeline---the tail-class scarcity proxy (\cref{fig:one-label-regime}) serves as the experimental counterpart to the label-evolution trigger.

\subsubsection{Token usage}
SEPDD's input tokens depend on the number of operators invoked and previous outputs; AIDE's are mainly determined by generated code and execution results. SEPDD's higher input usage arises from refinement loops and multi-LLM decomposition (separate LLMs for [idea generation, code creation] and [analysis, refinement], plus richer context). Since both share the same base model for code generation, AIDE produces fewer effective solutions per call.
Stronger base models reduce token usage: GPT-5.1's improved one-shot code generation and debugging substantially reduces operator invocations. Qwen-based SEPDD achieves slightly higher peak performance due to repeated runs, but for a comparable token budget GPT-based SEPDD has the potential for competitive or improved results.

\begin{table}[t]
    \centering
    \caption{Additional study.}
    \label{tab:addtional}

    \begin{subtable}{\linewidth}
        \centering
        \caption{Ablation study on \texttt{PVEL-AD} dataset. 
        }
        \label{tab:ablation}
        \begin{tabular}{lrrrr}
            \toprule
            Method       & Precision & Recall & mAP50 & mAP50-95 \\
            \midrule
            SEPDD \textit{(full)} & 90.0 & 85.0 & 91.5 & 62.1 \\
            SEPDD \textit{(greedy)} & 86.6 & 86.2 & 91.1 & 58.2 \\
            SEPDD \textit{w/o \revision{journal}} & 88.3 & 86.0 & 91.6 & 59.3 \\
            SEPDD \textit{(both)} & 85.2 & 84.9 & 91.4 & 55.0 \\
            SEPDD \textit{w/o refinement} & 87.5 & 83.2 & 90.1 & 65.0 \\
            \bottomrule
        \end{tabular}
    \end{subtable}

    \vspace{1em}

    \begin{subtable}{\linewidth}
        \centering
        \caption{Token usage.}
        \label{tab:toks}
        \begin{tabular}{llrrr}
            \toprule
            Dataset & Method       & Input & Output & Total \\
            \midrule
            \multirow{6}{*}{\texttt{PVEL-AD}} & AIDE & 0.88M & 0.19M & 1.06M \\
            & SEPDD (Qwen) & 2.73M & 0.21M & 2.94M \\
            & SEPDD (GPT-5.1) & 1.43M & 0.34M & 1.77M \\
            \cmidrule{2-5}
            & SEPDD \textit{(greedy)} & 1.56M & 0.31M & 1.88M \\
            & SEPDD \revision{\textit{w/o journal}} & 2.93M & 0.25M & 3.19M \\
            & SEPDD \textit{(both)} & 3.53M & 0.38M & 3.91M \\
            \midrule
            \multirow{3}{*}{\makecell{\tt \revision{EF-Industrial}}} & AIDE  & 1.36M & 0.23M & 1.59M \\
            & SEPDD (Qwen) & 1.27M & 0.23M & 1.50M \\
            & SEPDD (GPT-5.1) & 1.13M & 0.23M & 1.36M \\
            \bottomrule
        \end{tabular}
    \end{subtable}

\end{table}

\subsection{Multi-Round Adaptation Under 
Distribution Shift}
\label{sec:multiround}

To assess how
the framework behaves over \emph{consecutive} cycles and under a
\emph{continuously shifting} training distribution, we run
warm-started multi-round experiments on the {\tt PVEL-AD} and {\tt EF-Industrial} datasets.


\subsubsection{Protocol}
In both studies the \emph{test} split is fixed at
the full label space---eight classes on the {\tt PVEL-AD} dataset,
six on {\tt EF-Industrial}---so mAP50 is measured over the full label
space in every round. The \emph{training} labels, however, are
revealed incrementally: round~1 exposes only four common classes,
rounds~2 and~3 additionally unlock the remaining classes, and
rounds~4--5 hold the training distribution fixed (consolidation).
Each round is \emph{warm-started} from the previous round's best
solution, which is injected as the seed of the evolution journal, so
the search adapts an existing strong program rather than restarting
from scratch. This setup emulates a deployment in which new defect
categories arrive over time (continuous label-space shift) and the
model must integrate them without forgetting the classes already
mastered.

\revision{\subsubsection{Warm-start details (not from scratch)}
Only round~1 starts from
an empty journal (cold start, as in the \cref{tab:public,tab:industrial}, where all
autonomous comparisons also start from an empty root to keep search
budgets comparable); every subsequent round is warm-started. Formally,
at the beginning of round~$k>1$ the journal is initialized with a
single root node whose \texttt{code} field is the full Python pipeline
of round~$k-1$'s best node (retargeted to the current round's dataset
root); its metric is not carried over, so the root does not bias
node selection. The task description additionally tells the operators
that a strong prior solution is provided and that they should build on
it and focus adaptation on the newly unlocked class(es). No weights,
learning-rate schedules, or other training state are transferred---only
the pipeline code---because the search target is the training recipe
itself; cross-round reuse therefore happens at the program level, and
each round still trains detectors from scratch internally as part of
evaluation.}

\subsubsection{On the metric: seen-class control}
Because mAP50 is computed over the \emph{fixed} full-class label space,
a class with no training labels contributes an AP of zero by
construction, and one might ask whether the round~$1\!\rightarrow\!3$
rise is merely an artifact of the growing number of trainable classes.
To rule this out we re-evaluate every round's best model with a
\emph{single, uniform} validation pass and additionally report mAP50
restricted to the classes \emph{actually trained} that round
(\Cref{tab:multiround}
).

\begin{table}[t]
\centering
\caption{Warm-start multi-round evolution on the \revision{ public {\tt PVEL-AD} and private {\tt EF-Industrial} datasets}.}
\label{tab:multiround}
\setlength{\tabcolsep}{3.5pt}
\begin{tabular}{@{}l l c c c@{}}
\hline
Dataset & Round & Train classes & \multicolumn{1}{c}{mAP50} & mAP50 \\
        &       & (newly unlocked) & \multicolumn{1}{c}{(all classes)} & (trained only) \\
\hline
\multirow{5}{*}{\revision{{public}}} & \revision{1} & \revision{4 \ (cold start)} & \revision{43.8} & \revision{87.5} \\
                      & \revision{2} & \revision{6 \ (h.\,disloc., s.\,circuit)} & \revision{\textemdash} & \revision{\textemdash} \\
                      & \revision{3} & \revision{8 \ (star\_crack, v.\,disloc.)} & \revision{52.7} & \revision{52.7} \\
                      & \revision{4} & \revision{8 \ (consolidation)} & \revision{51.7} & \revision{51.7} \\
                      & \revision{5} & \revision{8 \ (consolidation)} & \revision{74.9} & \revision{74.9} \\
\hline
\multirow{5}{*}{{private}}    & 1 & 4 \ (cold start)         & 33.7  & 50.5 \\
                      & 2 & 5 \ (brokenGridLines)    & 40.5 & 48.6 \\
                      & 3 & 6 \ (blackSpots)         & 46.1 & 46.1 \\
                      & 4 & 6 \ (consolidation)      & 48.5 & 48.5 \\
                      & 5 & 6 \ (consolidation)      & 47.0 & 47.0 \\
\hline
\end{tabular}
\raggedright
\revision{\scriptsize\textit{Round~2 on {\tt PVEL-AD} produced no valid node (search failure) and reports no metric; its warm-start seed (round~1's best solution) is carried forward unchanged.}}
\end{table}

\subsubsection{PVEL-AD dataset}
\revision{Training labels are revealed incrementally by class
frequency: round~1 exposes the four most common classes
\{\texttt{black\_core}, \texttt{crack}, \texttt{finger},
\texttt{thick\_line}\}; round~2 additionally unlocks
\texttt{horizontal\_dislocation} and \texttt{short\_circuit}; round~3
unlocks the two rare tail classes \texttt{star\_crack} (121 training
instances) and \texttt{vertical\_dislocation} (112). 
}
\revision{The overall all-eight mAP50 rises from $43.8$ in the cold-start
round to $74.9$ by round~5 (\cref{tab:multiround}),
confirming that
warm-started evolution transfers useful structure on the public
dataset. The trajectory is, however, not monotone, and we
report it as measured. Round~2 is a search failure: none of its 19
nodes produced an executable pipeline (all failed at training or
evaluation with non-zero exit codes), so the round contributes no
metric and the warm-start chain carries round~1's best solution
forward unchanged into round~3. Round~3 therefore resumes from the
round-1 code with all eight classes unlocked and recovers immediately
($52.7$), and the consolidation rounds push the all-eight score to
$74.9$ with seven valid nodes in round~5. The seen-class control
(column ``trained only'') shows the expected pattern:
it starts high ($87.5$ on the four head classes), dips while the
harder mid- and tail-frequency classes are being absorbed ($52.7$,
$51.7$), and recovers during consolidation ($74.9$).
\emph{star\_crack}, the rarest class in {\tt PVEL-AD} (121 training
instances), is not recovered within any round on the public data (its
per-class mAP50 remains $0$ throughout, while the six unlocked
mid-frequency classes all reach usable APs), so single-round recovery
of classes at this rarity level appears to exceed what a one-round
search budget can deliver, and we do not claim it.}

\begin{figure}[t]
\centering
\includegraphics[width=0.89\linewidth]{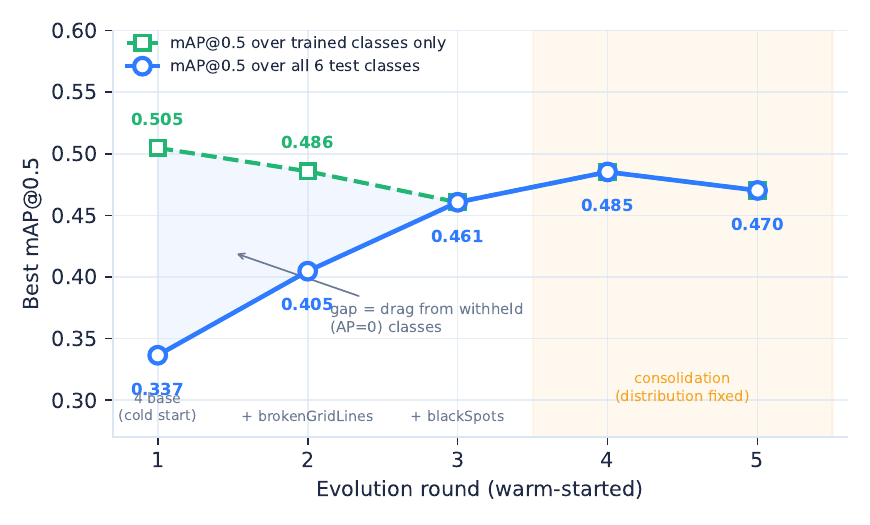}
\caption{\revision{Multi-round warm-start adaptation on the {\tt EF-Industrial} dataset. \emph{Blue}: best mAP50 over all six test classes; \emph{green (dashed)}: mAP50 over the classes actually trained that round. 
}}
\label{fig:multiround}
\end{figure}

\revision{\subsubsection{EF-Industrial dataset} The detector head is
fixed at $6$, and the \emph{training} labels are revealed
incrementally: round~1 exposes only the four common classes
$\{\text{microcracks}, \text{lightAndShade}, \text{blackSlice},
\text{blackEdging}\}$; round~2 additionally unlocks
\emph{brokenGridLines}; round~3 unlocks \emph{blackSpots}.}
\cref{tab:multiround} and \Cref{fig:multiround} report the
best mAP50 (over all six test classes) reached in each round. The
curve rises monotonically---$33.7 \rightarrow 40.5 \rightarrow
46.1$---as each newly introduced class is absorbed by the
warm-started search. The gains are concentrated on the classes unlocked
that round: a category withheld from training scores near zero and acts
as a fixed drag on the overall metric, and once it is introduced the
warm-started search recovers it within a single round while retaining
the previously learned classes. After all six classes are seen we run
two further consolidation rounds (4--5) with the training distribution
held fixed; performance saturates in the $46$--$49$ band (best
$0.485$ at round~4), confirming that the large gains stem from
absorbing newly introduced classes rather than from indefinite
re-optimization on an unchanged distribution. On {\tt EF-Industrial},
this seen-class score is high from the 
outset and stays essentially flat (green curve in
\cref{fig:multiround})---$50.5, 48.6, 46.1, 48.5,
47.0$ across rounds~1--5---whereas the all-six score climbs from
$33.7$ to meet it; the mild dip reflects the newly added classes
being \emph{harder} (\emph{brokenGridLines} $\sim\!39$,
\emph{blackSpots} $\sim\!40$), not degradation of earlier ones,
which hold throughout (e.g.\ \emph{microcracks} $83$--$85$). The
improvement comes from \emph{absorbing new classes} without
catastrophic forgetting, not from re-normalizing the metric. This confirms that the
framework adapts to continuous distribution shift and
warm-starting transfers useful structure across rounds rather than
discarding it.
The PVEL-AD trajectory is less
smooth than 
EF-Industrial 
: its round~2 search failure and the
rounds~3--4 dip are absent on EF-Industrial, whose all-six curve rises
monotonically as each newly unlocked class is absorbed within a single
round. The tail classes differ as well---the rarest EF-Industrial
classes are roughly ten times more frequent than \emph{star\_crack}
and are recovered within one round, whereas
\emph{star\_crack} is not, so single-round recovery appears limited to
classes above this rarity level.

\revision{\subsubsection{Computational cost}
Beyond tokens (\cref{tab:toks}), \cref{tab:cost} reports wall-clock and GPU cost. With four parallel workers, a single PVEL-AD cycle has a median wall-clock of 12.4\,h (26.0 GPU-hours summed per node, 155 LLM calls); the EF-Industrial multi-round protocol is substantially cheaper (6.6\,h wall / 7.9 GPU-hours / 282 calls across five rounds) because it trains on 814 images against PVEL-AD's 4{,}050. Per valid node, training/evaluation takes roughly 15--25 GPU-minutes on PVEL-AD and 5--10 on EF-Industrial. Cost scales linearly with the search budget ($k \times c \times T$ nodes), as analyzed in \cref{sec:limitations}; invalid nodes---typically 2--9 of 19 per run---waste on the order of 10--20\% of the GPU budget, which the per-node refinement loop keeps bounded. We report the two datasets separately because their per-node protocols differ (the PVEL-AD runs use the full inspector/refiner debug loop, the EF-Industrial multi-round runs use the single-pass \texttt{simple\_evolve} protocol).}

\begin{table}[t]
  \centering
  \caption{\revision{Computational cost per evolution cycle.
  }}
  \label{tab:cost}
  \setlength{\tabcolsep}{4pt}
  \begin{tabular*}{\linewidth}{@{\extracolsep{\fill}}@{} l l r r r @{}}
    \toprule
    \multicolumn{1}{l}{\revision{Dataset}} &
    \multicolumn{1}{l}{\revision{Setting}} &
    \multicolumn{1}{r}{\revision{Wall (h)}} &
    \multicolumn{1}{r}{\revision{GPU-h}} &
    \multicolumn{1}{r}{\revision{LLM calls}} \\
    \midrule
    \revision{\texttt{PVEL-AD}} &
      \makecell[l]{\revision{Single cycle}\\\revision{(10-run median)}} &
      \revision{12.4} & \revision{26.0} & \revision{155} \\
    \midrule
    \multirow{2}{*}{\revision{\texttt{EF-Industrial}}} &
      \revision{5 rounds (total)} & \revision{6.6} & \revision{7.9} & \revision{282} \\
    & \revision{per round (avg.)} & \revision{1.3} & \revision{1.6} & \revision{56} \\
    \bottomrule
  \end{tabular*}
\end{table}
\subsection{Limitations and Failure Modes}
\label{sec:limitations}

While the proposed framework attains strong detection
accuracy without manual model engineering, it inherits several
limitations that are intrinsic to LLM-driven program search. We
discuss these together with the observed failure modes, so
practitioners can anticipate when the method degrades.

\subsubsection{Reliance on the underlying LLM}
Every candidate solution is proposed, implemented, and refined by the
backbone LLM; consequently the search ceiling is bounded by the
model's coding and reasoning ability. A weaker backbone produces a
larger fraction of syntactically or semantically invalid programs,
shrinking the effective search width for a fixed budget. The
framework is therefore not self-contained: its performance is coupled
to, and improves with, the capability of the language model it is
instantiated on.

\subsubsection{Invalid search branches}
Because candidate code is generated rather than hand-written, a
non-trivial fraction of nodes fail validation or execution and must be
pruned. \revision{Figure~\ref{fig:evolution-tree}} shows one cycle in which
node~11 produced an invalid solution that yielded no usable metric and
was discarded by the top-$k$ selection, so search budget spent on that
branch returns no accuracy gain. Ablating the code-refinement operator
makes this worse: without automatic repair, far fewer generated nodes
survive to evaluation, leaving the journal too sparse for effective
best-first expansion. The top-$k$ mechanism contains the damage---an
invalid node cannot enter the active set---but it cannot recover the
wasted compute.

\revision{\subsubsection{False-trigger risk} 
False triggering is not catastrophic but has a cost. Wasted cost is bounded because each trigger merely starts an \emph{offline} search cycle, whose outcome is gated: candidates are promoted only if they beat the currently deployed model on the fixed holdout (\cref{sec:operational-impl}), so a false trigger typically wastes one cycle of tokens and GPU time (on the order of \cref{tab:toks} and \cref{tab:cost}) rather than degrading the deployed detector. Journal pollution is bounded by versioning: a cycle triggered on stale data appends nodes under a run tag that can be excluded from warm-start seeds, and the promoted-model pointer simply remains at the previous artifact. Finally, because promotion is validated and rollback is supported, the residual risk of a false trigger reduces to compute expense, which the co-firing rule and calibrated thresholds keep rare.
}

\subsubsection{Degradation under extreme distribution shift}
When a defect category is entirely absent from the training labels,
the evolved model cannot detect it regardless of search quality, and
its per-class AP collapses. The multi-round
experiment (\cref{fig:multiround}) makes this explicit: in the early rounds the
tail classes are withheld from training and their AP is near zero;
once each class is introduced, warm-started search recovers it within
a round. The framework adapts to \emph{incremental} shift but offers
no remedy for classes it has never seen---an inherent limitation of
the supervised detector it evolves, not of the search itself.

\subsubsection{Computational cost of evolution cycles}
A single cycle issues on the order of
$k \times c \times T$ LLM completions (with top-$k$ width $k$,
$c$ children per node, and $T$ iterations) and trains an equal number
of full detectors for evaluation. Cost thus scales linearly with the
search budget in both API tokens and GPU-hours; \revision{\cref{tab:toks,tab:cost}
report the measured token, wall-clock, and GPU cost on our setup}. This makes very wide or very deep searches expensive, and is
the main obstacle to running many consecutive cycles in practice.

The failures above are graceful rather than catastrophic: invalid
branches are pruned, and unseen classes score low instead of
corrupting 
model. Nonetheless they define the regime
in which the framework is effective---a capable backbone LLM, a search
budget commensurate with 
available compute, and distribution shift that is incremental rather than abrupt.

\section{Conclusion} \label{sec:conclusion}

This paper presented \textit{SEPDD}, a self-evolving photovoltaic defect detection framework for industrial electroluminescence inspection, addressing limited and long-tailed training data, complex defect morphology, distribution shifts, and emerging defect categories. By combining automated pipeline optimization with a trigger-driven evolution mechanism that re-optimizes the full training/evaluation pipeline at deployment time, the framework enables robust, adaptive detection with minimal human intervention. Experiments on a public benchmark and a private industrial dataset showed that SEPDD consistently improves detection of reliability-critical and low-contrast defects under severe class imbalance and domain shift, adapting more effectively to evolving categories than fixed pipelines and manually tuned baselines. Future work will extend the framework to multimodal inspection and fleet-level PV reliability management.

\section*{Acknowledgements}

The authors acknowledge the use of artificial intelligence (AI) tools for minor language editing and revision of this manuscript. The authors have reviewed and validated all content and take full responsibility for the final version.


\begin{footnotesize}
\bibliographystyle{IEEEtran}
\bibliography{mybib}
    
\end{footnotesize}

\end{document}